%% file: main.tex
\documentclass{article}

\PassOptionsToPackage{numbers, compress}{natbib}

\usepackage[main, preprint]{neurips_2026}

\usepackage[utf8]{inputenc} 
\usepackage[T1]{fontenc}    
\usepackage{hyperref}       
\usepackage{url}            
\usepackage{booktabs}       
\usepackage{amsfonts}       
\usepackage{nicefrac}       
\usepackage{microtype}      
\usepackage{xcolor}         

\usepackage{lipsum}
\usepackage{duckuments}
\usepackage{graphicx}
\usepackage{subcaption}
\usepackage{booktabs}
\usepackage[table]{xcolor}
\usepackage{amssymb}
\usepackage{multirow}
\usepackage{amsmath}
\usepackage{pdflscape}      
\usepackage{adjustbox}      
\usepackage{multirow}       
\usepackage[table]{xcolor}  
\usepackage{array}          

\usepackage[most]{tcolorbox}
 \usepackage[table]{xcolor}
 \usepackage{booktabs}
 \usepackage{multirow}
 \usepackage{colortbl}
\usepackage{wrapfig}
\usepackage{pifont}
\newcommand{\xmark}{\ding{55}} 
\definecolor{heavyBg}{HTML}{FDEDEC}   
\definecolor{midBg}{HTML}{FEF9E7}     
\definecolor{lightBg}{HTML}{E8F8F5}   
\definecolor{ourRow}{HTML}{AED6F1}    
\definecolor{headerCol}{HTML}{1A5276} 
\definecolor{groupRule}{HTML}{B2BABB} 

\newtcolorbox{researchquestion}{
  colback=gray!15,
    colframe=gray!40,
  boxrule=1pt,
  arc=1pt,
  left=10pt, right=10pt, top=6pt, bottom=6pt,
  halign=center,
  fontupper=\itshape,
  enhanced,
  breakable
}

\hypersetup{
  colorlinks   = true, 
  urlcolor     = blue, 
  linkcolor    = red, 
  citecolor   = blue 
}

\usepackage{graphicx}
\usepackage{adjustbox}

\title{PERL: Parameter Efficient Reasoning in CLIP Latent Space}

\author{%
  Simone Carnemolla \textsuperscript{$\dagger$} \\
  University of Catania\\
  \texttt{simone.carnemolla@phd.unict.it} \\
  \And
   Salvatore Calcagno \textsuperscript{$\dagger$}\\
   University of Catania \\
  \texttt{salvatore.calcagno@phd.unict.it} \\
  \And
   Daniela Giordano \\
   University of Catania \\
  \texttt{daniela.giordano@unict.it} \\
   \And
  Concetto Spampinato \\
   University of Catania \\
  \texttt{concetto.spampinato@unict.it} \\
  \And
  Matteo Pennisi \\
   University of Catania \\
  \texttt{matteo.pennisi@unict.it} \\
}

\begin{document}

\maketitle

\begingroup
\renewcommand{\thefootnote}{\fnsymbol{footnote}}
\footnotetext[2]{Equal Contribution}
\endgroup

\begin{abstract}
\input{sections/0_abstract}
\end{abstract}

\section{Introduction}
\input{sections/1_intro_new}

\section{Related Work}
\input{sections/2_related_new}

\section{Method}
\input{sections/3_method}
\section{Experimental Results}
\input{sections/4_results}

\section{Conclusion}
\input{sections/5_conclusion}




\newpage
\bibliographystyle{plainnat}
\bibliography{biblio}


\clearpage
\newpage
\appendix

\section*{Technical appendices and supplementary material}

\section{Implementation Details}
\label{sec:impl-details}
\input{sections/appendix_A_implementation_details}

\clearpage
\newpage
\section{Extended Base to Novel Evaluation}
\label{sec:b2n_extended}
\input{sections/appendix_B_b2n_extended}

\clearpage
\newpage
\section{Relation to weight-shared iterative architectures.}
\label{sec:relation-to-iterative}
\input{sections/appendix_C_technical_supplementary}

\clearpage
\newpage
\section{Comparison with Baselines in Original Setting}
\label{sec:original_setting}
\input{sections/appendix_D_original_baselines}

\clearpage
\newpage
\section{Out of Distribution Evaluation}
\label{sec:ood}
\input{sections/appendix_E_ood}

\clearpage
\newpage
\section{Extended Analysis for number of Reasoning Steps}
\label{sec:extended_num_thoughts_analysis}
\input{sections/appendix_F_extended_analysis}


\end{document}

%% file: sections/0_abstract.tex
Contrastively trained vision-language models such as CLIP provide strong zero-shot transfer by aligning images and text in a shared embedding space. However, adapting these models to downstream tasks without degrading their open-vocabulary generalization remains challenging. Existing parameter-efficient adaptation methods typically improve task specialization through learned prompts, adapters, or multimodal transformations, where adaptation capacity is primarily expressed through additional trainable parameters. Inspired by recent latent reasoning methods in language models, we investigate a complementary perspective: \emph{can adaptation emerge from iterative reasoning on latent representations rather than from increasing parameter count alone?} We introduce \textbf{PERL} (Parameter-Efficient Reasoning in CLIP Latent Space), a lightweight adaptation framework that augments a frozen CLIP model with a compact shared reasoning module applied recurrently across refinement steps. At each step, PERL generates a latent reasoning token conditioned on the current representation and injects it into an intermediate encoder layer, progressively refining higher-level semantic representations while preserving CLIP's pretrained multimodal structure. Across 15 benchmarks spanning base-to-novel generalization, cross-dataset transfer, and out-of-distribution ImageNet variants, PERL achieves the best parameter-performance trade-off among the compared methods under a fast-adaptation few-shot setting, combining strong novel-class accuracy and competitive transfer performance with only about 6K trainable parameters, up to 817x fewer than the largest compared approach.
Overall, our results suggest that iterative latent reasoning provides a complementary adaptation mechanism to parameter scaling in discriminative vision-language models.

%% file: sections/1_intro_new.tex
Adapting CLIP-style vision-language models~\cite{clip} to downstream tasks is often framed as adding a small amount of task-specific flexibility while keeping the pretrained image and text encoders fixed. This constraint is attractive because it preserves CLIP's broad image-text alignment and zero-shot transfer ability, while enabling lightweight downstream specialization. A large body of work explores this trade-off through learned prompts, multimodal prompt transformations, visual perturbations, low-rank updates, cache-based mechanisms, or adapter modules. Prompt tuning methods learn context vectors~\cite{coop}; conditional prompting methods such as CoCoOp generate instance-dependent prompts~\cite{cocoop}; multimodal prompting methods introduce coordinated visual-language transformations~\cite{maple}; and adapter-based approaches insert compact trainable modules inside the backbone~\cite{gao2024clip, houlsby2019, lora}. These approaches have delivered strong gains and are largely complementary. Despite their architectural differences, however, many express adaptation primarily through additional learned parameters, support statistics, or optimization-dependent components.

Recent latent reasoning methods in language models suggest a complementary perspective. Approaches such as Coconut~\cite{coconut} and Quiet-STaR~\cite{quietstar} show that additional capability can emerge not only from increasing model size, but also from iteratively refining hidden representations over multiple computation steps. In these systems, the model effectively ``thinks longer'' by repeatedly updating its latent state before prediction. While latent reasoning has been studied primarily in autoregressive language models, the broader principle is more general: repeated computation can provide expressivity complementary to parameter scaling.

In this work, we investigate whether this principle transfers to frozen discriminative vision-language models. Rather than increasing adaptation capacity primarily through larger or more specialized trainable modules, we ask whether a small shared module can gain effective capacity by being applied recurrently over the model's evolving latent state. Our central hypothesis is that, in a frozen vision-language model, adaptation need not arise only from storing more task-specific parameters; it can also emerge from progressively steering pretrained representations toward task-relevant decision regions while remaining compatible with CLIP's original multimodal embedding space.

This perspective is especially relevant for few-shot and rapidly deployable specialization of frozen foundation models. In practical settings, a single backbone may support many downstream adaptations across users, domains, tasks, or deployment environments. The cost of adaptation is therefore not only inference latency, but also the cumulative burden of training, storing, selecting, and maintaining many specialized parameter sets. In such settings, trainable parameters must remain small, while a modest increase in per-example computation is often acceptable.

\begin{wrapfigure}{r}{0.40\linewidth}

\centering
\includegraphics[width=0.39\textwidth]{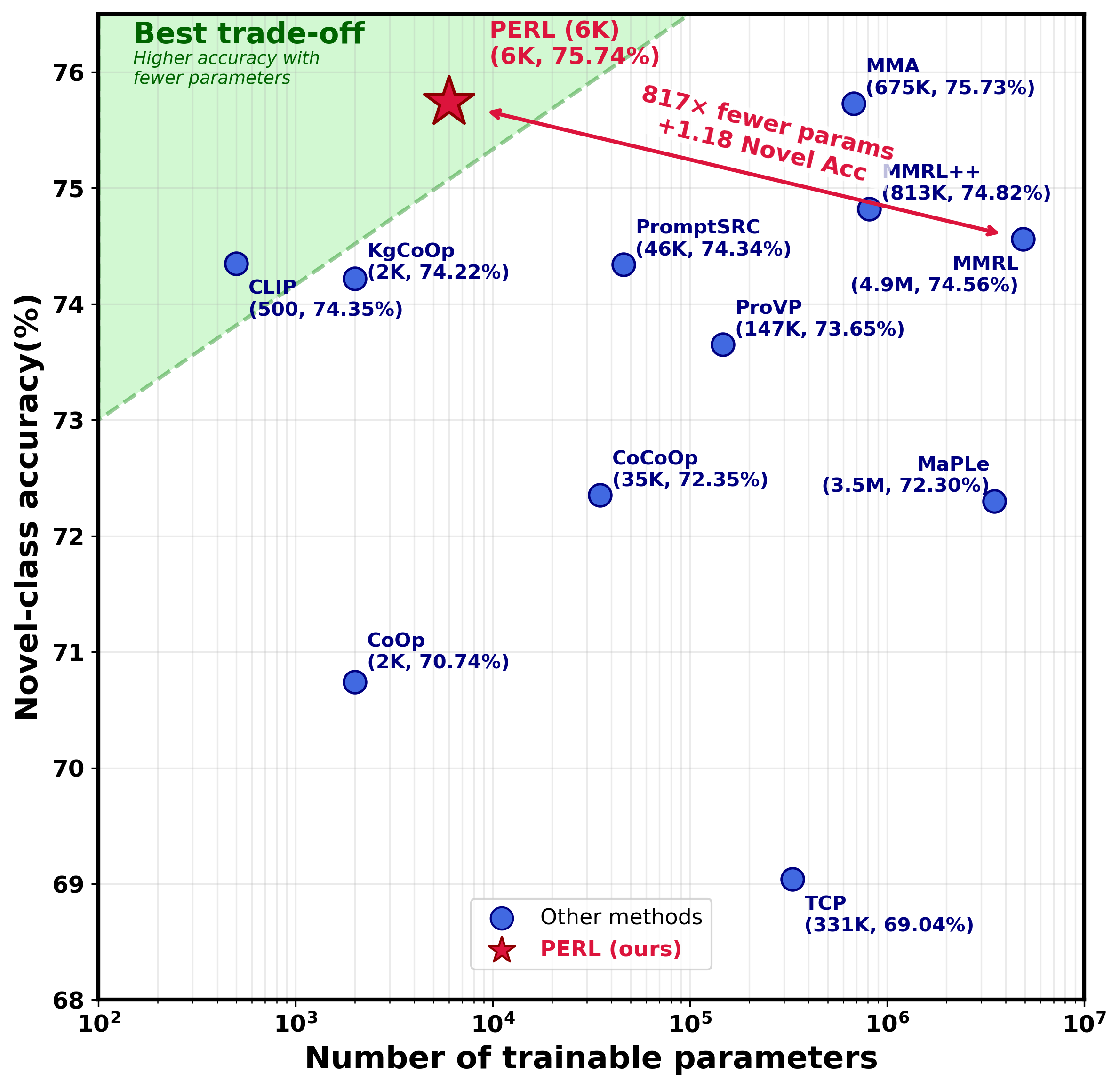}
\caption{
\textbf{Novel-class accuracy versus trainable parameters for representative CLIP adaptation methods.}
PERL achieves competitive novel-class generalization with only 6K trainable parameters, suggesting that iterative latent reasoning can complement parameter scaling in CLIP adaptation.}
\label{fig:teaser}

\end{wrapfigure}

We instantiate this idea in \textbf{PERL} (\emph{Parameter-Efficient Reasoning in CLIP Latent Space}), a lightweight adaptation framework in which a compact shared module recurrently refines latent representations over multiple computation steps. Rather than increasing adaptation capacity through additional task-specific parameters, PERL reuses the same small reasoning module iteratively, allowing the model to trade a modest increase in computation for improved task specialization while keeping the CLIP backbone frozen and avoiding test-time gradient updates.

Figure~\ref{fig:teaser} situates PERL within the parameter-performance trade-off of representative CLIP adaptation methods. PERL achieves novel-class generalization competitive with substantially larger adaptation frameworks while operating with only 6K trainable parameters, placing it in a favorable low-parameter regime among the compared methods. These results suggest that iterative latent reasoning can complement parameter scaling as a mechanism for adaptation capacity.

At the same time, transferring latent reasoning to CLIP is nontrivial. Unlike autoregressive language models, CLIP operates through contrastive alignment between image and text embeddings, and preserving access to this geometry is central to maintaining generalization. Iterative refinement in this setting should therefore not freely rewrite representations, but instead reshape them in a controlled manner that remains compatible with CLIP's pretrained multimodal structure.

We therefore situate our investigation in a constrained few-shot adaptation regime and ask:

\begin{researchquestion}
\textit{
Can latent reasoning on internal representations provide an adaptation axis complementary to parameter scaling in frozen discriminative vision-language models?
}
\end{researchquestion}

PERL is designed around three principles. First, \emph{weight sharing across refinement steps} lets effective adaptation capacity grow with computation rather than with parameter count. Second, \emph{modality-specific reasoning}. Third, \emph{intermediate-layer insertion} allows refinement to influence semantic representations while limiting perturbations to lower-level pretrained features, helping retain CLIP's pretrained transfer behavior.

These design choices naturally motivate evaluation under constrained optimization budgets. A single-epoch few-shot setting reflects deployments where support data, tuning time, and per-task maintenance are limited, while also isolating how much adaptation can be expressed architecturally rather than accumulated through repeated optimization. In this regime, iterative latent reasoning exposes a different trade-off from conventional parameter scaling: a small learned module can be reused multiple times at inference to provide additional computation without storing additional per-step parameters.

Empirically, we conduct an extensive evaluation across 15 benchmarks covering base-to-novel generalization, cross-dataset transfer, and out-of-distribution ImageNet variants. Despite using a fully frozen CLIP backbone and only 6K trainable parameters, PERL outperforms common compact prompt-tuning baselines and approaches substantially larger multimodal adaptation methods. Overall, the results indicate that iterative latent reasoning can provide meaningful adaptation capacity even in extremely parameter-constrained regimes.

\noindent\textbf{Contributions:}
\begin{itemize}
    \item \textbf{Computation as an adaptation resource.}
    We investigate iterative latent reasoning as an adaptation mechanism for frozen contrastive vision-language models, showing that repeated computation can provide useful task-specific refinement at fixed parameter count.

    \item \textbf{PERL: a recurrent lightweight adaptation mechanism for CLIP.}
    We introduce a frozen-backbone adaptation framework in which a compact shared projector recurrently refines latent representations through dynamically generated reasoning tokens.

    \item \textbf{Strong parameter-performance trade-offs under constrained adaptation.}
    Under a single-epoch few-shot adaptation protocol, PERL improves over common prompt-tuning baselines on base-to-novel generalization, remains competitive on cross-dataset transfer and out-of-distribution evaluation, and approaches substantially larger multimodal adaptation methods using orders of magnitude fewer parameters.
\end{itemize}

%% file: sections/2_related_new.tex
A large body of work adapts pre-trained CLIP models~\cite{clip} to downstream tasks while keeping the backbone frozen. Prompt-learning methods such as CoOp~\cite{coop} replace the hand-crafted \textit{``a photo of a [class]''} template with learnable continuous context tokens, while CoCoOp~\cite{cocoop} conditions these prompts on the input image to improve generalization beyond base classes. Later methods extend this idea with multimodal prompts~\cite{maple}, regularization toward zero-shot CLIP representations~\cite{promptsrc}, knowledge-guided initialization~\cite{kgcoop}, progressive visual prompting~\cite{provp}, decoupled base--novel objectives~\cite{dept}, and class-aware textual prompts~\cite{tcp}. Adapter-based methods such as MMA~\cite{mma} introduce lightweight trainable modules inside the frozen encoders. Closest to our setting are MMRL~\cite{mmrl} and MMRL++~\cite{mmrl++}, which introduce modality-agnostic representation tokens at intermediate layers of both CLIP encoders to encourage shared vision-language adaptation while preserving the frozen backbone. These methods establish the value of shared latent mediation for multimodal adaptation; PERL builds on this direction but asks whether a lightweight latent mechanism can be reused recurrently, so that additional adaptation capacity comes from iterative computation rather than a larger set of learned tokens or modules.

A complementary line of work investigates iterative latent computation through repeated refinement of hidden representations. Weight-shared iterative architectures, including Universal Transformers~\cite{universaltransformer}, deep equilibrium models~\cite{deq}, and looped transformers~\cite{loopedtransformers}, increase effective computation by repeatedly applying the same transformation rather than adding a separate set of parameters at each depth. More recently, language-model work such as Coconut~\cite{coconut} and Ouro~\cite{ouro} has explored latent or looped computation as a way to increase effective reasoning depth without proportional parameter growth, while Quiet-STaR and Fast Quiet-STaR~\cite{quietstar, fastquietstar} study learned internal rationales before token prediction. These results motivate iterative latent computation, but they target autoregressive language modeling rather than frozen discriminative vision-language adaptation.

PERL lies at the intersection of these directions: parameter-efficient CLIP adaptation and shared-weight iterative latent computation. It does not operate autoregressively or generate intermediate textual reasoning traces. Instead, it applies shared-weight latent refinement inside a frozen contrastive vision-language model, using recurrent updates to refine multimodal representations under tight parameter budgets. To our knowledge, prior CLIP adaptation work has not studied shared-weight recurrent latent refinement as the adaptation mechanism with frozen contrastive vision-language backbones. This distinguishes PERL from MMRL/MMRL++, which use shared learned representation tokens but do not perform iterative refinement with a reused update module (see also Appendix~\ref{sec:relation-to-iterative}).

%% file: sections/3_method.tex
\begin{figure}[t]
\centering
\includegraphics[width=0.85\textwidth]{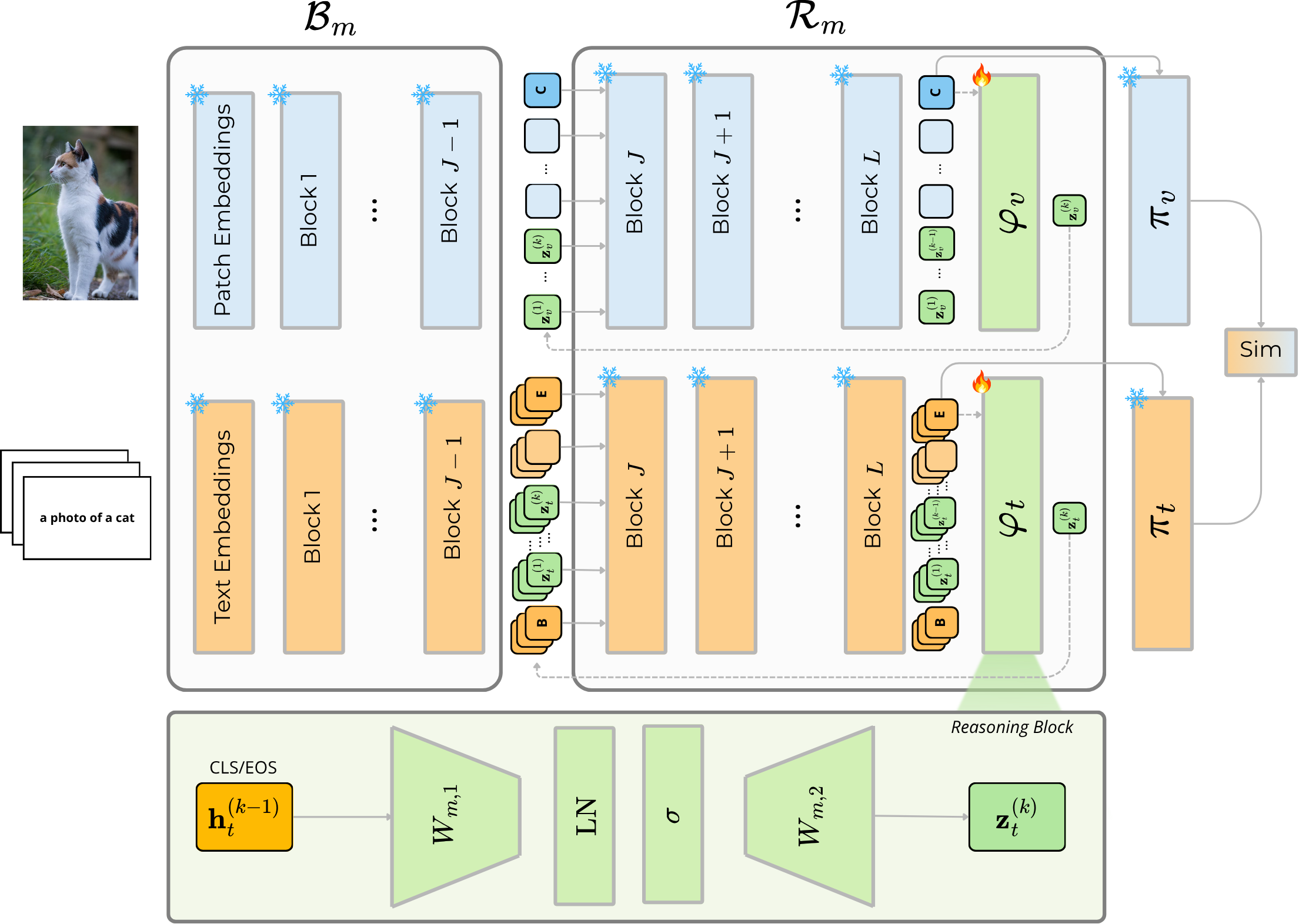}
\caption{\textbf{PERL architecture.} PERL adds iterative latent reasoning to a frozen CLIP model. Each encoder is split into frozen base $\mathcal{B}_m$ and refinement $\mathcal{R}_m$ modules; a small trainable projector $\varphi_m$, shared across steps, recurrently maps the pooled state into a reasoning token $\mathbf{z}_m^{(k)}$ appended before the refinement blocks. This trades extra compute for adaptation capacity without adding trainable parameters. Refined CLS/EOS are projected by the frozen heads $\pi_v, \pi_t$ and matched via similarity.}
\label{fig:method}
\end{figure}

\paragraph{Preliminaries.}
Figure~\ref{fig:method} illustrates the overall PERL architecture. Let $\mathcal{V}$ and $\mathcal{T}$ denote the frozen CLIP vision and text encoders, and let $\pi_v$ and $\pi_t$ denote the corresponding projection heads. Each encoder contains $L$ Transformer blocks, with hidden dimensions $d_v$ and $d_t$, respectively. The projection heads
$\pi_v : \mathbb{R}^{d_v} \rightarrow \mathbb{R}^{d}$
and
$\pi_t : \mathbb{R}^{d_t} \rightarrow \mathbb{R}^{d}$
map pooled modality-specific features to CLIP's shared $d$-dimensional embedding space. All CLIP parameters are kept frozen during training.

\paragraph{Encoder split.}
For each modality $m \in \{v,t\}$, we split the frozen encoder at depth $J$ into a base module $\mathcal{B}_m$, containing the lower $J$ Transformer blocks, and a refinement module $\mathcal{R}_m$, containing the remaining $L-J$ blocks. Given an input token sequence $\mathbf{x}_m$, we first compute the frozen base representation as $\mathbf{S}_m = \mathcal{B}_m(\mathbf{x}_m)$. The zero-shot pooled representation is then obtained as $\mathbf{h}_{m}^{(0)} = \mathrm{Pool}_m(\mathcal{R}_m(\mathbf{S}_m))$, where $\mathrm{Pool}_v$ reads the CLS token for images and $\mathrm{Pool}_t$ reads the EOS token for text.

\paragraph{PERL.}
For each modality $m$, we introduce a lightweight thought projector
$\varphi_m : \mathbb{R}^{d_m} \rightarrow \mathbb{R}^{d_m}$, implemented as a two-layer MLP with a rank-$r$ bottleneck preceded by Layer Normalization. The projector is defined as
\begin{equation}
    \varphi_m(\mathbf{h})
    =
    W_{m,2}
    \sigma
    \left(
        W_{m,1}\mathrm{LN}(\mathbf{h})
    \right),
\end{equation}
where $W_{m,1} \in \mathbb{R}^{r \times d_m}$, $W_{m,2} \in \mathbb{R}^{d_m \times r}$, and $\sigma$ denotes a non-linear activation function. These projectors are the only trainable components of the model.

Starting from the zero-shot pooled representation $\mathbf{h}_m^{(0)}$, each modality performs $K$ refinement steps. At iteration $k$, the projector generates a thought token from the previous pooled representation:
\begin{equation}
    \mathbf{z}_m^{(k)}
    =
    \varphi_m
    \left(
        \mathbf{h}_m^{(k-1)}
    \right),
    \qquad
    k = 1,\dots,K .
\end{equation}
The newly generated token is appended to the frozen base sequence together with all previously generated thought tokens. The augmented sequence is then processed by the refinement module:
\begin{equation}
\label{eq:perl-recursion}
    \mathbf{h}_m^{(k)}
    =
    \mathrm{Pool}_m
    \left(
        \mathcal{R}_m
        \left(
            \mathbf{z}_m^{(1)}
            \Vert
            \dots
            \Vert
            \mathbf{z}_m^{(k)}
            \Vert
            \mathbf{S}_m
        \right)
    \right).
\end{equation}

Here, $\Vert$ denotes concatenation along the sequence dimension. At each step $k$, the prior thought tokens $\mathbf{z}_m^{(1)},\dots,\mathbf{z}_m^{(k-1)}$ are re-included in the input sequence;
the refinement module $\mathcal{R}_m$ is therefore called $K{+}1$ times per modality (once for the zero-shot pooled state $\mathbf{h}_m^{(0)}$, then once per refinement step). The same projector $\varphi_m$ is reused across all $K$ refinement steps within each modality, so the refinement dynamics remain fixed across steps: capacity grows with the number of recurrent applications rather than with a per-step parameter set, and no step-specific transformation is learned.

\paragraph{Readout invariance.} A central design choice in PERL is that the pooled representation is always read from the original CLS (image) or EOS (text) position, never from a thought-token position. Thought tokens participate in self-attention as ordinary tokens---contributing to the keys and values that shape the readout---but never serve as the readout themselves. 
The value of the CLS/EOS state is still influenced by the thought tokens through self-attention.

\paragraph{Prediction and training objective.}
After $K$ refinement steps, we project the final pooled representations using the $\pi_v$ and $\pi_t$ frozen projection heads, obtaining the refined image embedding $\mathbf{v}_i^{(K)}$ and the refined text embeddings $\mathbf{T}^{(K)} \in \mathbb{R}^{C \times d}$ over the $C$ candidate classes. The corresponding zero-shot embeddings, $\mathbf{v}_i^{(0)}$ and $\mathbf{t}_j^{(0)}$, are obtained from the same frozen model with $K=0$.

Given a training example $(\mathbf{x}_i,y_i)$, with $y_i\in\{1,\dots,C\}$, the logits $\boldsymbol{\ell}_i$ are computed as $\tau \mathbf{v}_i^{(K)}(\mathbf{T}^{(K)})^\top$, where $\tau$ is CLIP's frozen logit scale. The primary objective is the standard cross-entropy loss over the candidate classes. To preserve CLIP's zero-shot geometry, we additionally penalize deviations between the refined and zero-shot image embeddings in cosine geometry. The resulting training objective is:
\begin{equation}
    \mathcal{L}
    =
    \underbrace{
    \frac{1}{N}
    \sum_{i=1}^{N}
    \mathcal{L}_{\mathrm{CE}}
    \left(
        \boldsymbol{\ell}_i,
        y_i
    \right)
    }_{\mathcal{L}_{\mathrm{cls}}}
    +
    \lambda
    \underbrace{
        \left(
        1
        -
        \frac{1}{N}
        \sum_{i=1}^{N}
        \cos
        \left(
            \mathbf{v}_i^{(0)},
            \mathbf{v}_i^{(K)}
        \right)
        \right)
    }_{\mathcal{L}_{\mathrm{anchor}}} .
\end{equation}

Here, $\mathcal{L}_{\mathrm{cls}}$  rewards refinements that improve task-specific separability, while $\mathcal{L}_{\mathrm{anchor}}$ penalizes departures from the zero-shot manifold that are not justified by classification gain. The resulting equilibrium is a refinement that adjusts the pooled image representation only as much as the downstream task requires---precisely the inductive bias we want in fast adaptation, where overfitting to a small support set is the dominant failure mode. We anchor only the visual embedding because, in CLIP, the textual classifier is shared across all examples in a batch, so its drift is already constrained by every classification update.

\input{tabs/base2new}

%% file: sections/4_results.tex
\subsection{Datasets}
We evaluate our approach on fifteen image-classification benchmarks covering
generic object recognition, fine-grained classification, scene understanding,
texture recognition, satellite imagery, action recognition, and robustness to
distribution shift. Following ~\citet{coop} protocol, we use
\textbf{ImageNet}~\citep{imagenet},
\textbf{Caltech101}~\citep{caltech101},
\textbf{OxfordPets}~\citep{oxfordpets},
\textbf{StanfordCars}~\citep{stanfordcars},
\textbf{Flowers102}~\citep{oxfordflowers},
\textbf{Food101}~\citep{food101},
\textbf{FGVC-Aircraft}~\citep{fgvcaircraft},
\textbf{SUN397}~\citep{sun397},
\textbf{DTD}~\citep{dtd},
\textbf{EuroSAT}~\citep{eurosat}, and
\textbf{UCF101}~\citep{ucf101}.
These datasets provide a broad testbed, with category counts ranging from $10$
classes in EuroSAT to $1000$ classes in ImageNet. Additionally results about distribution shift between ImageNet
and its variants:
\textbf{ImageNet-V2}~\citep{imagenetv2},
\textbf{ImageNet-Sketch}~\citep{imagenetsketch},
\textbf{ImageNet-A}~\citep{imageneta}, and
\textbf{ImageNet-R}~\citep{imagenetr} are reported in Appendix ~\ref{sec:ood}.

\subsection{Experimental Setup}
\label{sec:exp_set}

We consider two complementary evaluation protocols:
\textbf{-- Base-to-novel generalization}: the classes of each dataset are split evenly
into base and novel categories. The model is trained only on the base classes and
evaluated on both base and novel classes. This protocol measures whether the
method can improve task-specific adaptation on seen classes while preserving the
zero-shot generalization ability of the pre-trained vision-language model on
unseen categories. We report accuracy on base and novel classes, as well as their
harmonic mean.

\textbf{-- Cross-dataset transfer}: we train the model on all $1000$ ImageNet classes in
the few-shot setting and directly evaluate it on the remaining downstream
datasets without any additional fine-tuning. This setting assesses whether the
learned adaptation transfers beyond the source dataset. Extensive experiments on domain generalization across Imagenet variants are reported in Appendix \ref{sec:ood}.

All experiments use CLIP with a ViT-B/16 visual backbone. PERL is optimized with AdamW using a learning rate of $10^{-4}$, batch size $4$, and $K=4$ reasoning steps. For both base-to-novel and cross-dataset evaluations, PERL and all baselines are trained for a single epoch in the 16-shot setting (\emph{fast adaptation}), emphasizing parameter-efficient specialization under limited optimization. Further implementation details and results under the original baseline training settings are reported in Appendix~\ref{sec:impl-details} and Appendix~\ref{sec:original_setting}.

\subsection{Results}

\paragraph{Base-to-novel.}
Table~\ref{tab:base2new_1} shows that PERL is particularly effective in the base-to-novel setting, achieving an average HM of $74.79$ with only 6K trainable parameters. 
PERL performs particularly well on novel classes, attaining the highest average novel accuracy within numerical precision ($75.74$; MMA: $75.73$), while outperforming larger methods such as MMRL++ and PromptSRC. On the overall harmonic mean across base and novel classes, PERL remains within $0.44$ points of MMA and $0.87$ of PromptSRC while using much fewer trainable parameters ($113\times$ less than MMA and $8\times$ than PromptSRC).
Overall, the results suggest that iterative latent reasoning improves task adaptation while preserving CLIP's transfer ability to unseen categories (see also Appendix ~\ref{sec:b2n_extended}).
\begin{table}[ht!]

\definecolor{highlightCol}{HTML}{FEF9E7}

\caption{\textbf{Base-to-novel generalization.} Models are trained on base classes and evaluated on both base and novel classes; we report the harmonic mean (HM) on each dataset. \textbf{\#Params} reports trainable parameters; \textbf{$\times$PERL} the ratio to PERL's parameter count; \textbf{$\Delta$HM} the average-HM gap vs.\ PERL 
}
\label{tab:base2new_1}
\centering

\setlength{\tabcolsep}{3pt}
\renewcommand{\arraystretch}{1.4}
\setlength{\aboverulesep}{0pt}
\setlength{\belowrulesep}{0pt}

\resizebox{\textwidth}{!}{%
\begin{tabular}{@{}l r >{\columncolor{highlightCol}}r ccccccccccc ccc >{\columncolor{highlightCol}}r@{}}
\toprule
\textbf{Method} & \textbf{\#Params} & \textbf{$\times$PERL}
& \multicolumn{11}{c}{\textbf{Datasets (HM)}}
& \multicolumn{4}{c}{\textbf{Average}} \\
\cmidrule(lr){4-14}\cmidrule(lr){15-18}
 & & & 
\rotatebox{45}{ImageNet} & 
\rotatebox{45}{Caltech101} & 
\rotatebox{45}{OxfordPets} & 
\rotatebox{45}{StanfordCars} & 
\rotatebox{45}{Flowers102} & 
\rotatebox{45}{Food101} & 
\rotatebox{45}{FGVCAircraft} & 
\rotatebox{45}{SUN397} & 
\rotatebox{45}{DTD} & 
\rotatebox{45}{EuroSAT} & 
\rotatebox{45}{UCF101} 
& Base & Novel & HM & $\Delta$HM \\
\midrule
CLIP$_{\text{ICML'21}}$ & 0 & --- & 70.22 & 95.40 & 94.12 & 68.65 & 74.83 & 90.66 & 31.09 & 72.23 & 56.99 & 60.03 & 73.85 & 69.34 & 74.35 & 71.64 & $-3.15$ \\
\midrule
MMRL$_{\text{CVPR'25}}$    & 4.9M & 817$\times$ & 73.81 & 94.77 & 92.06 & 68.86 & 73.43 & 89.99 & 31.99 & 75.06 & 56.02 & 66.66 & 74.27 & 70.92 & 74.27 & 72.45 & $-2.34$\\
MMRL++$_{\text{IJCV'25}}$  & 813K & 136$\times$ & 73.81 & 95.66 & 92.83 & 69.75 & 74.74 & 89.69 & 32.22 & 75.08 & 55.83 & 69.10 & 75.05 & 71.81 & 74.50 & 73.07 & $-1.72$ \\
MaPLe$_{\text{CVPR'23}}$   & 3.5M & 583$\times$ & 73.47 & 94.69 & 96.29 & 70.60 & 80.33 & 91.15 & 14.19 & 77.88 & 64.06 & 71.72 & 77.27 & 75.69 & 72.30 & 73.79 & $-1.00$ \\
MMA$_{\text{CVPR'24}}$     & 675K & 113$\times$ & 73.60 & 95.69 & 94.92 & 72.17 & 83.08 & 91.06 & 34.67 & 78.76 & 65.92 & 58.42 & 79.26 & 75.18 & 75.73 & 75.23 & $+0.44$ \\
TCP$_{\text{CVPR'24}}$     & 331K & 55$\times$  & 66.75 & 89.42 & 88.53 & 65.40 & 68.02 & 89.14 & 20.77 & 64.63 & 45.93 & 64.55 & 71.61 & 64.85 & 69.04 & 66.80 & $-7.99$ \\
ProVP$_{\text{IJCV'24}}$   & 147K & 25$\times$  & 71.65 & 95.13 & 94.34 & 70.24 & 75.66 & 90.59 & 32.33 & 74.93 & 61.96 & 57.16 & 76.33 & 72.00 & 73.65 & 72.76 & $-2.03$ \\
CoCoOp$_{\text{CVPR'22}}$    & 35K & 6$\times$  & 72.97 & 95.33 & 95.91 & 70.47 & 78.41 & 90.46 & 24.14 & 76.65 & 62.50 & 64.22 & 76.63 & 74.90 & 72.35 & 73.43 & $-1.36$ \\
PromptSRC$_{\text{ICCV'23}}$ & 46K  & 8$\times$  & 73.19 & 96.27 & 96.20 & 71.37 & 82.00 & 91.04 & 33.57 & 78.78 & 58.90 & 71.56 & 79.37 & 77.45 & 74.34 & 75.66 & $+0.87$ \\
KgCoOp$_{\text{CVPR'23}}$ & 2K & 0.33$\times$ & 58.41 & 95.68 & 92.98 & 68.90 & 72.87 & 90.07 & 30.11 & 72.77 & 56.86 & 58.41 & 71.49 & 66.36 & 74.22 & 69.87 & $-4.92$ \\
CoOp$_{\text{IJCV'22}}$   & 2K & 0.33$\times$ & 72.86 & 95.24 & 93.29 & 69.85 & 75.58 & 89.88 & 20.49 & 76.66 & 56.64 & 52.03 & 74.86 & 70.73 & 70.74 & 70.67 & $-4.12$ \\
\midrule
\textbf{PERL (Ours)}      & 6K & 1$\times$    & 72.95 & 96.22 & 96.11 & 70.38 & 77.59 & 91.13 & 33.08 & 76.97 & 60.87 & 69.62 & 77.78 & 73.97 & 75.74 & 74.79 & --- \\
\bottomrule
\end{tabular}}
\end{table}

\paragraph{Cross-dataset.}

Table~\ref{tab:cross_dataset_1} reports transfer from ImageNet to ten downstream datasets after a single epoch of few-shot adaptation. PERL achieves an average target accuracy of $65.56$, outperforming all compact prompt-tuning baselines and remaining competitive with substantially larger multimodal adaptation methods. In particular, PERL is within $0.13$ points of MaPLe despite using $583\times$ fewer trainable parameters, and within two points of MMRL++ while using $136\times$ fewer parameters. PERL also achieves the best performance on StanfordCars and UCF101.

Compared to the stronger gains in the base-to-novel setting, these results suggest that PERL primarily preserves and refines CLIP's pretrained transfer geometry rather than over-specializing to the source dataset, yielding a favorable parameter-transfer trade-off under a very small trainable footprint.

\begin{table}[h!]

\definecolor{highlightCol}{HTML}{FEF9E7}  

\caption{\textbf{Cross-dataset evaluation.} Models are trained on ImageNet (source) and evaluated zero-shot on 10 target datasets. \textbf{\#Params} reports trainable parameters; \textbf{$\times$PERL} the ratio to PERL's count; \textbf{$\Delta$Avg.} the gap in mean target accuracy vs.\ PERL 
}
\label{tab:cross_dataset_1}
\centering
\scriptsize
\setlength{\tabcolsep}{3pt}
\renewcommand{\arraystretch}{1.4}
\setlength{\aboverulesep}{0pt}
\setlength{\belowrulesep}{0pt}

\resizebox{\textwidth}{!}{%
\begin{tabular}{@{}l r >{\columncolor{highlightCol}}r c cccccccccc c >{\columncolor{highlightCol}}r@{}}
\toprule
\textbf{Method} & \textbf{\#Params} & \textbf{$\times$PERL} & \textbf{Source} & \multicolumn{10}{c}{\textbf{Target}} & \textbf{Average} & \textbf{$\Delta$Avg.} \\
\cmidrule(lr){4-4}\cmidrule(lr){5-14}

 & & & 
\rotatebox{45}{ImageNet} & 
\rotatebox{45}{Caltech101} & 
\rotatebox{45}{OxfordPets} & 
\rotatebox{45}{StanfordCars} & 
\rotatebox{45}{Flowers102} & 
\rotatebox{45}{Food101} & 
\rotatebox{45}{FGVCAircraft} & 
\rotatebox{45}{SUN397} & 
\rotatebox{45}{DTD} & 
\rotatebox{45}{EuroSAT} & 
\rotatebox{45}{UCF101} & & \\
\midrule
MMRL$_{\text{CVPR'25}}$    & 4.9M & 817$\times$ & 72.03 & 94.67 & 91.43 & 66.10 & 72.77 & 86.40 & 26.30 & 67.57 & 45.90 & 53.10 & 68.27 & 67.25 & $+1.69$ \\
MMRL++$_{\text{IJCV'25}}$  & 813K & 136$\times$ & 71.87 & 94.63 & 91.43 & 66.60 & 73.53 & 86.73 & 26.07 & 67.77 & 46.13 & 53.00 & 69.03 & 67.49 & $+1.93$ \\
MaPLe$_{\text{CVPR'23}}$   & 3.5M & 583$\times$ & 70.53 & 93.93 & 90.10 & 65.30 & 72.17 & 86.20 & 23.17 & 66.90 & 46.03 & 45.37 & 67.73 & 65.69 & $+0.13$ \\
MMA$_{\text{CVPR'24}}$     & 675K  & 113$\times $        & 71.33 & 93.57 & 90.20 & 66.30 & 71.67 & 86.20 & 24.97 & 67.90 & 44.37 & 36.30 & 68.65 & 65.01 & $-0.55$ \\
TCP$_{\text{CVPR'24}}$     & 331K & 55$\times$  & 65.23 & 89.46 & 86.10 & 62.83 & 65.56 & 85.46 & 19.26 & 59.30 & 41.20 & 45.60 & 66.03 & 62.08 & $-3.48$ \\
ProVP$_{\text{IJCV'24}}$   & 147K & 25$\times$  & 68.90 & 93.17 & 90.57 & 64.85 & 69.15 & 85.71 & 24.52 & 65.94 & 46.02 & 43.24 & 67.68 & 65.09 & $-0.47$ \\
CoCoOp$_{\text{CVPR'22}}$    & 35K & 6$\times$ & 70.40 & 93.63 & 90.00 & 66.53 & 71.83 & 86.17 & 23.83 & 67.10 & 45.90 & 41.30 & 68.70 & 65.50 & $-0.06$ \\
PromptSRC$_{\text{ICCV'23}}$ & 46K  & 8$\times$       & 70.30 & 93.67 & 89.83 & 65.96 & 71.16 & 86.33 & 23.83 & 66.76 & 47.23 & 44.19 & 58.50 & 64.75 & $-0.81$ \\
KgCoOp$_{\text{CVPR'23}}$ & 2K & 0.33$\times$ & 67.38 & 93.51 & 88.74 & 65.78 & 68.06 & 85.58 & 23.68 & 63.45 & 45.07 & 41.18 & 65.71 & 64.08 & $-1.48$ \\
CoOp$_{\text{IJCV'22}}$   & 2K & 0.33$\times$ & 70.30 & 93.63 & 89.90 & 65.13 & 70.93 & 85.96 & 22.90 & 65.94 & 46.00 & 46.83 & 67.06 & 65.50 & $-0.06$ \\
\midrule
\textbf{PERL (Ours)}      & 6K & 1$\times$    & 70.25 & 94.33 & 90.36 & 66.92 & 70.79 & 86.37 & 24.60 & 67.46 & 45.72 & 40.01 & 69.05 & 65.56 & --- \\
\bottomrule
\end{tabular}}
\end{table}

\subsection{Ablation}

\input{tabs/ablation}

\paragraph{Modality ablation.}
We first analyze which modality benefits most from reasoning by comparing text-only, vision-only, and joint text-vision variants of PERL (Table~\ref{tab:ablation_v_t_b}). The joint configuration performs best, indicating that reasoning is most effective when it can simultaneously refine textual class representations and visual evidence. Among the single-modality variants, text-only reasoning yields stronger gains than vision-only reasoning, consistent with the central role of language supervision in CLIP adaptation~\cite{coop, cocoop, kgcoop}. However, the gap between text-only and joint reasoning shows that refining the textual decision space alone is insufficient; reasoning over visual representations provides complementary information.

\begin{figure}
    \centering

    \begin{subfigure}{0.48\linewidth}
        \centering
        \includegraphics[width=\linewidth]{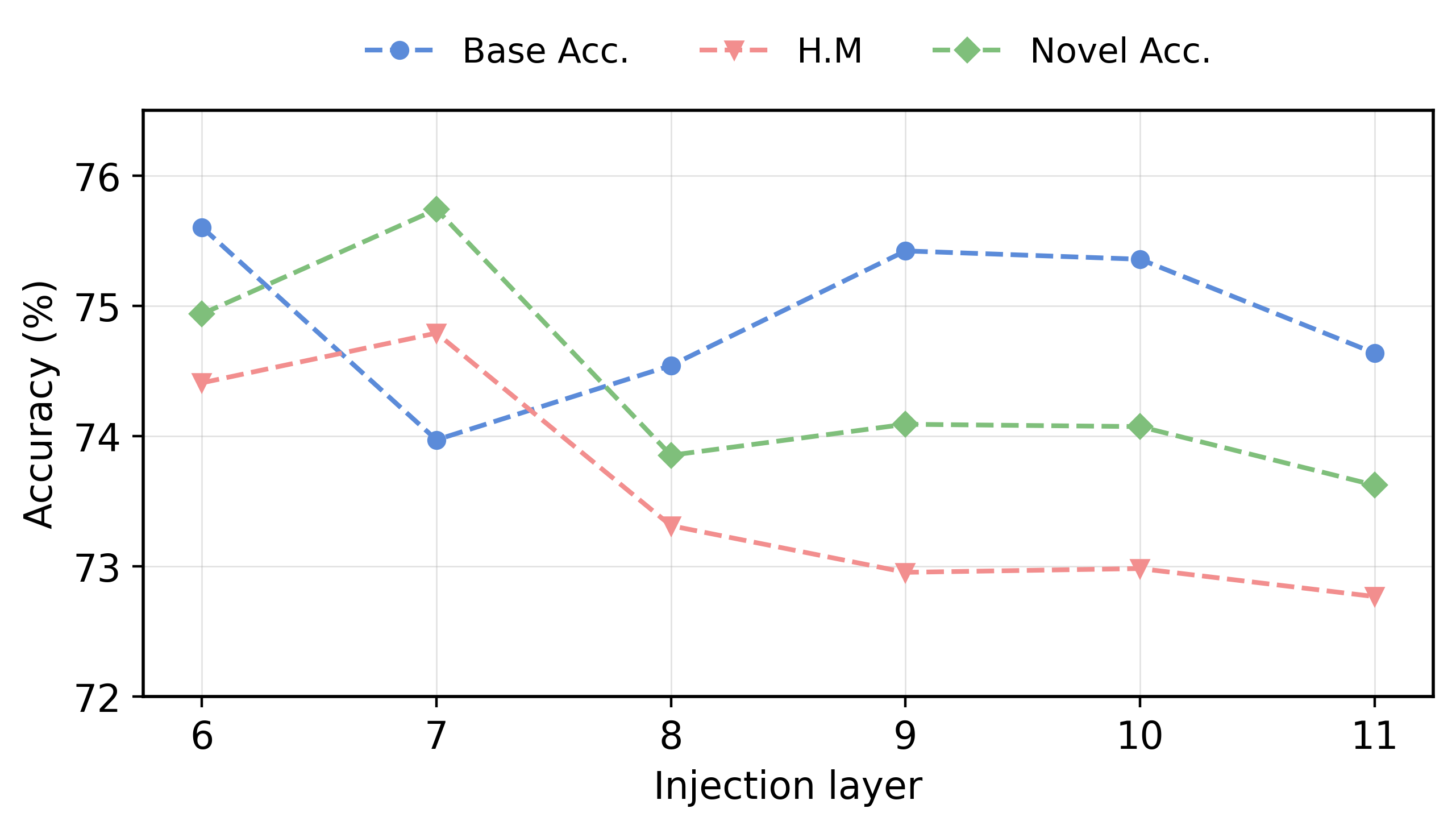}
        \caption{Injection layer}
        \label{fig:injection_layer}
    \end{subfigure}
    \hfill
    \begin{subfigure}{0.48\linewidth}
        \centering
        \includegraphics[width=\linewidth]{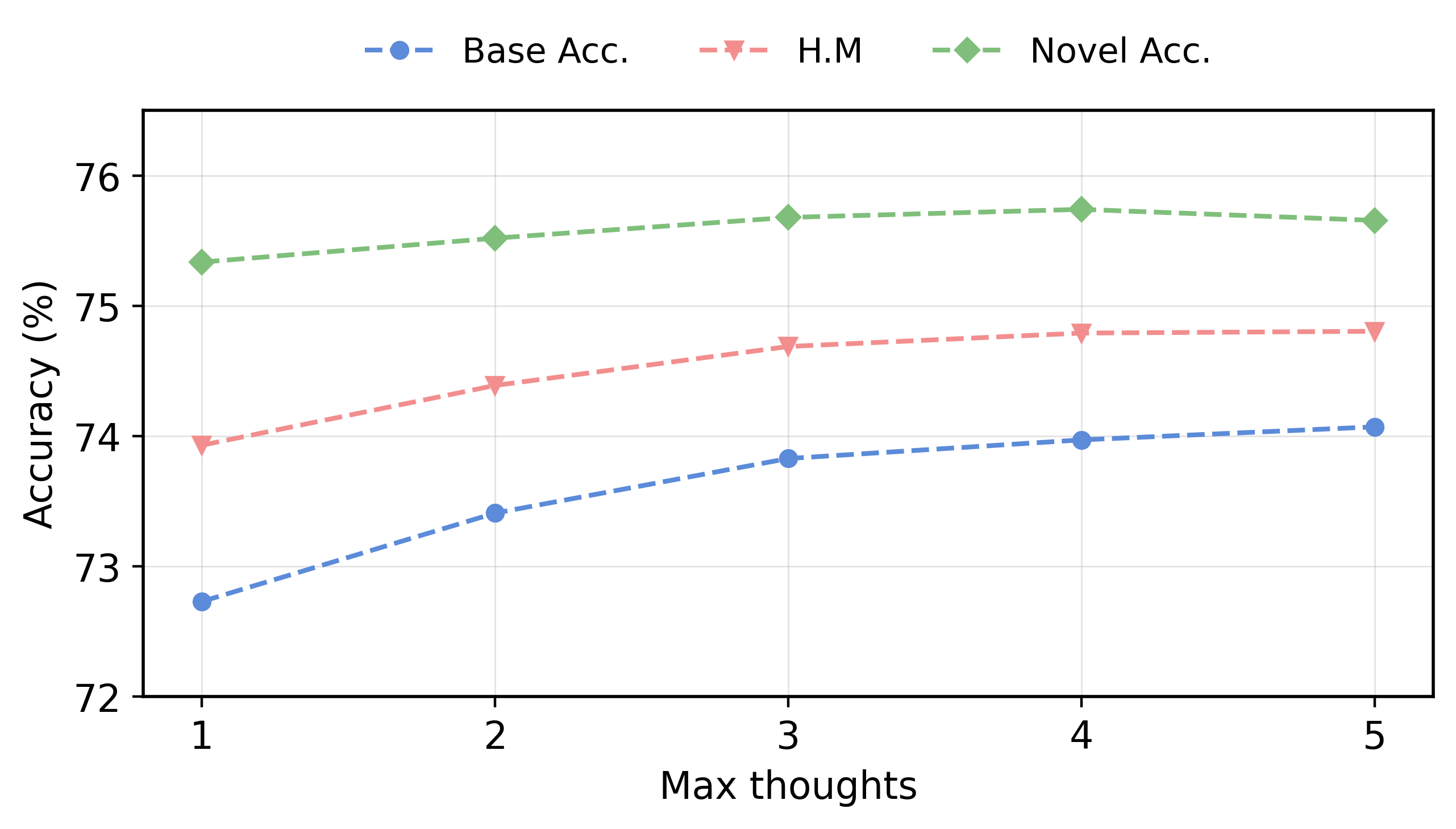}
        \caption{Reasoning steps}
        \label{fig:max_thoughts}
    \end{subfigure}

    \caption{\textbf{Ablation on reasoning configurations.} Base, Novel, and Harmonic Mean accuracy w.r.t the injection layer (left) and number of reasoning steps (right). Performance peaks at layer 7 with 4 reasoning iterations.}
    \label{fig:ablation}
\end{figure}

\paragraph{Reasoning Configuration Analysis.}
We first analyze how reasoning should be distributed across the transformer hierarchy (Table~\ref{tab:ablation_reasoning}). PERL-Early injects reasoning at the first encoder layer, PERL-Shared (default) applies a shared reasoning module only at an intermediate semantic layer, and PERL-Deep injects reasoning at multiple depths. The results reveal a clear trend: early reasoning substantially degrades performance, indicating that perturbing low-level visual features is detrimental. In contrast, restricting reasoning to intermediate semantic layers provides the best trade-off between adaptation and generalization, as also observed in MMRL~\citep{mmrl}, while multi-depth reasoning improves base accuracy at the cost of novel-class transferability. Motivated by this observation, we next analyze which semantic layer is most suitable for reasoning. Fig.~\ref{fig:ablation} (top) evaluates reasoning-token injection from layers 6 to 11 of the visual encoder. Performance peaks at layer 7, which achieves the best harmonic mean and highest novel accuracy. As reasoning moves toward deeper layers, base accuracy increases while novel accuracy progressively decreases, again reflecting a specialization–transfer trade-off.
\begin{table}[t]
\centering
\scriptsize
\setlength{\tabcolsep}{4pt}
\renewcommand{\arraystretch}{1.2}

\caption{\textbf{Ablation of reasoning-loop configurations} on base-to-novel generalization. $J$ denotes the encoder layers receiving reasoning-token injections. \textit{Skip-early} restricts reasoning to intermediate/deep layers, while \textit{Multi-depth} applies reasoning at multiple depths. \textit{Layer-} and \textit{Step-specific} indicate whether separate reasoning projectors are used across layers or reasoning iterations.}

\label{tab:ablation_reasoning}
\begin{tabular}{@{}l r lcccc ccc@{}}
\toprule
\textbf{Variant} & \textbf{\#Params}
 &\textbf{J}& \textbf{Skip early}
& \textbf{Multi-depth}
& \textbf{Layer-specific}
& \textbf{Step-specific}
& \textbf{Base} & \textbf{Novel} & \textbf{HM} \\
\midrule
PERL-Early & 6K   &0&    \xmark      &    \xmark         & \xmark        &     \xmark        & 68.91          & 62.43          & 65.15          \\
PERL-Deep & 6K   &[3, 7, 11]&  \xmark           & \checkmark &      \xmark       &     \xmark        & \textbf{75.70} & 73.66          & 74.52          \\
\textbf{PERL-Shared (default)}              & \textbf{6K}  &7& \checkmark &   \xmark          & \xmark        &      \xmark       & 73.97 & \textbf{75.74} & \textbf{74.79} \\
\midrule
PERL-Step                & 32K&7& \checkmark &    \xmark         & \xmark         & \checkmark & 74.30          & 75.28          & 74.93 \\
PERL-Layer              & 26K& [7, 9, 11]& \checkmark & \checkmark & \checkmark &      \xmark       & 75.58            & 74.83            & 75.09            \\
PERL-LayerStep         & 83K&[7, 9, 11]& \checkmark & \checkmark & \checkmark & \checkmark & \textbf{77.17}           &  \textbf{75.74 }           & \textbf{76.37}            \\

\bottomrule
\end{tabular}
\end{table}

Fig.~\ref{fig:ablation} (bottom) analyzes the number of reasoning iterations. Performance improves steadily from one to four reasoning steps, after which gains saturate, suggesting that iterative latent reasoning progressively refines predictions rather than applying a single corrective update.

\paragraph{Reasoning Capacity Analysis.} Finally, we investigate whether reasoning provides a capacity axis complementary to parameter scaling. The default PERL configuration shares the same reasoning projector across all layers and reasoning steps, resulting in only 6K trainable parameters. We then progressively increase reasoning capacity by introducing independent projectors across layers  (PERL-Layer), reasoning steps  (PERL-Step), or both (PERL-LayerStep). Table~\ref{tab:ablation_reasoning} shows that performance improves consistently as additional reasoning modules are introduced, with the strongest configuration reaching a harmonic mean of 76.37 using only 83K trainable parameters. These results suggest that iterative reasoning itself provides meaningful adaptation capacity, which can further scale when additional reasoning modules are allocated.

\subsection{Dynamics of Iterative Reasoning}
\label{sec:dynamics}

To better understand the effect of iterative reasoning beyond final accuracy, we analyze how predictions and visual evidence evolve across reasoning steps. We consider four complementary metrics averaged across datasets: accuracy, target confidence, Brier score, and the Jacobian norm of the target logit with respect to the input.

\begin{figure}[t]
    \centering
     \begin{minipage}[t]{0.48\textwidth}
        \centering
        \includegraphics[width=0.99\textwidth]{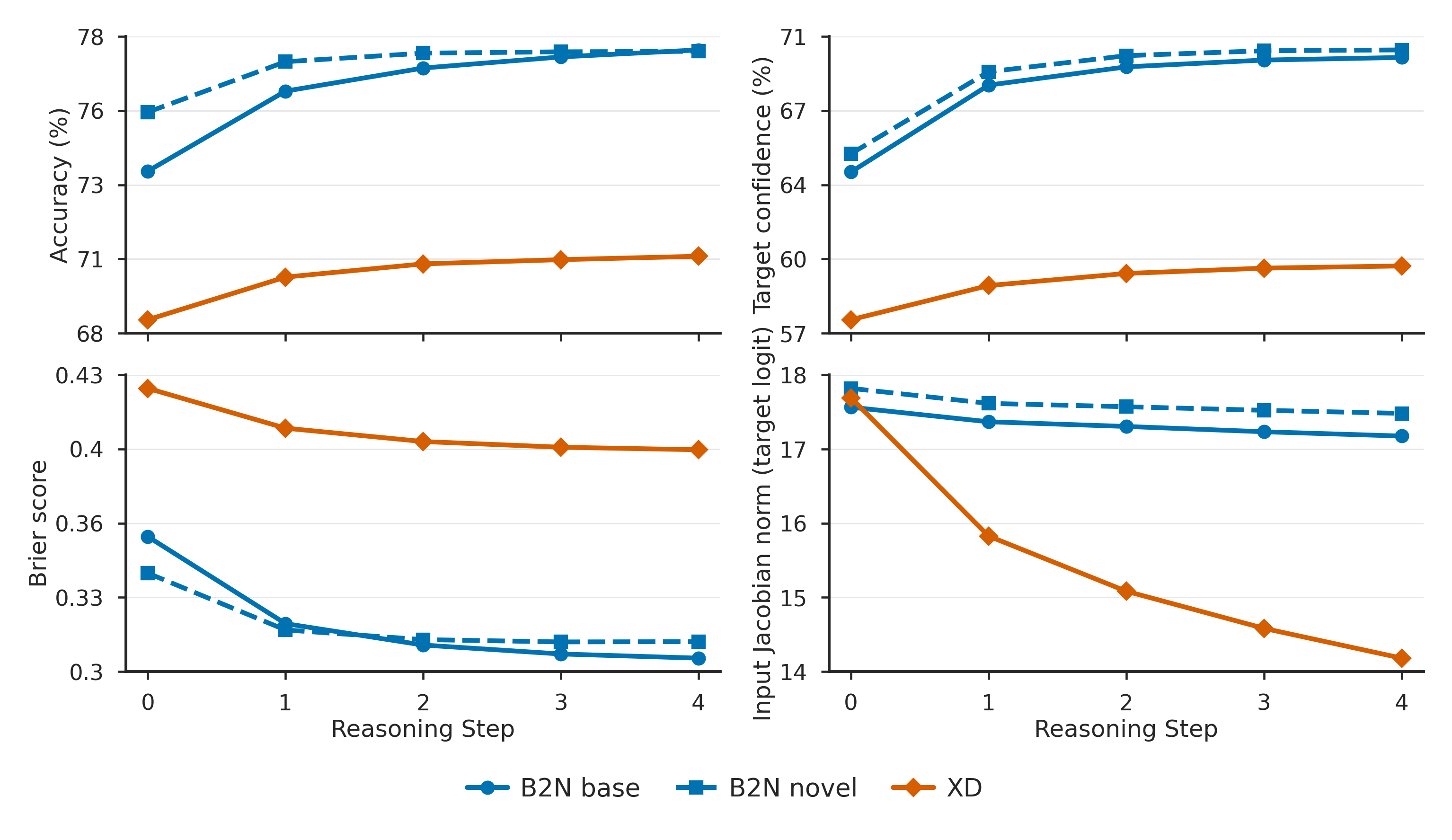}
        \caption{\textbf{Dynamic Reasoning Metrics.} Accuracy, target confidence, Brier score, and Jacobian norm of the target logit w.r.t.\ the input, evaluated in the base-to-novel and cross-dataset settings.}
        \label{fig:accuracy-target-conf-brier-input-jacobian-target}
    \end{minipage}
    \hfill
    \begin{minipage}[t]{0.48\textwidth}
        \centering
        \includegraphics[width=0.6\textwidth]{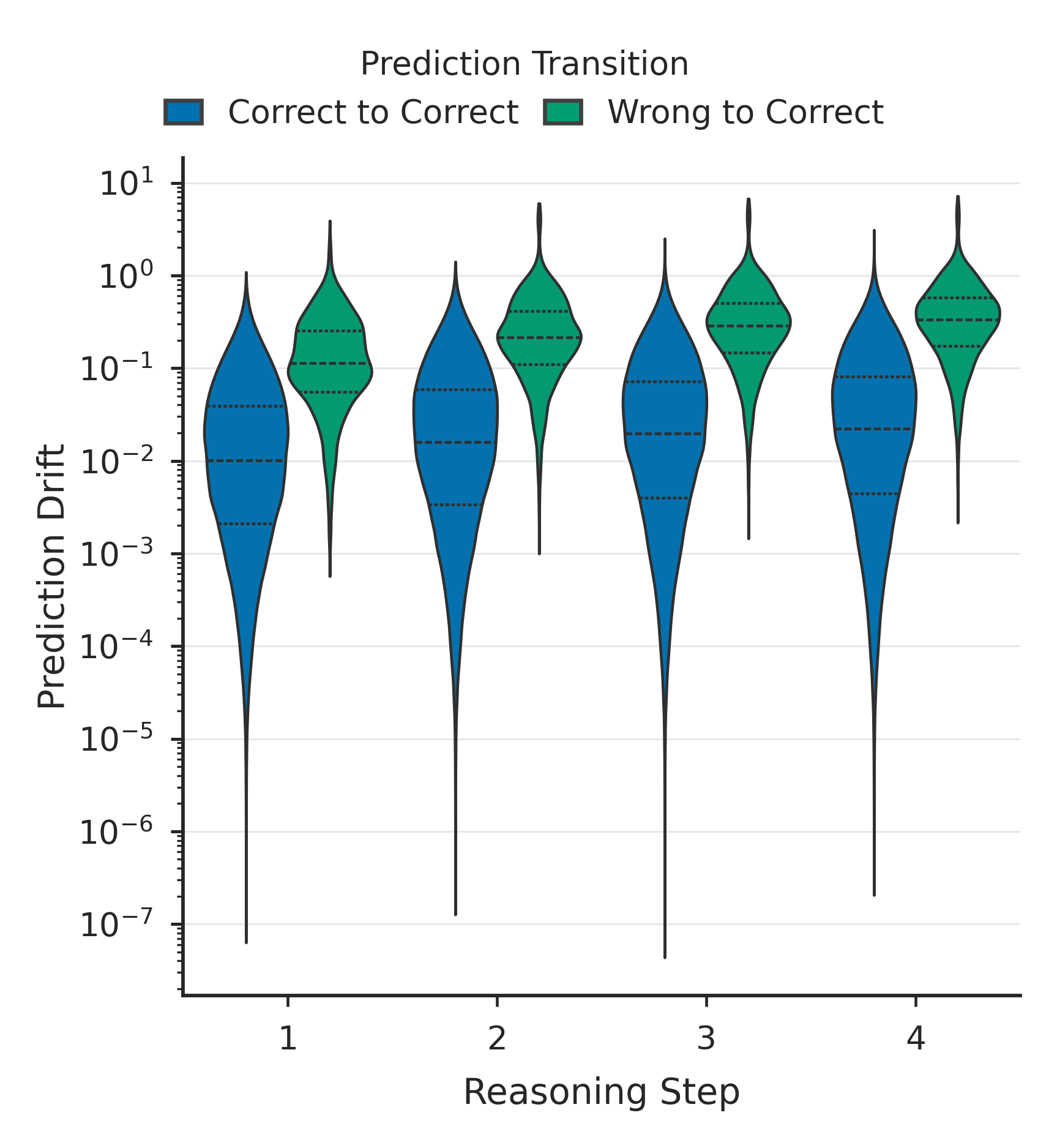}
        \caption{\textbf{Per-sample prediction drift.} KL divergence from the step-0 distribution at each reasoning step, grouped by setting and transition type (\textit{Correct-to-Correct}, \textit{Wrong-to-Correct}). Larger values indicate stronger drift.}
        \label{fig:kl_to_step0}
    \end{minipage}
\end{figure}

Fig.~\ref{fig:accuracy-target-conf-brier-input-jacobian-target} reveals a progressive refinement process across reasoning steps. Most gains occur at the first iteration, where accuracy, confidence, and calibration improve substantially, indicating that reasoning rapidly resolves ambiguities in the initial prediction. Later iterations provide smaller but consistent gains, suggesting that iterative reasoning progressively refines the representation rather than applying a single corrective update. At the same time, the decreasing Jacobian norm indicates that predictions become progressively more stable and less sensitive to input perturbations.

We next analyze how much the predictive distribution changes during reasoning. Fig.~\ref{fig:kl_to_step0} compares the KL divergence between each reasoning step and the initial step-0 prediction. Samples that are already correct exhibit only limited drift, whereas wrong-to-correct transitions produce progressively larger changes. PERL therefore does not apply a uniform transformation to all samples, but instead performs stronger updates when the initial prediction requires correction.

\begin{figure}[htbp]
    \centering
    \includegraphics[width=0.95\textwidth]{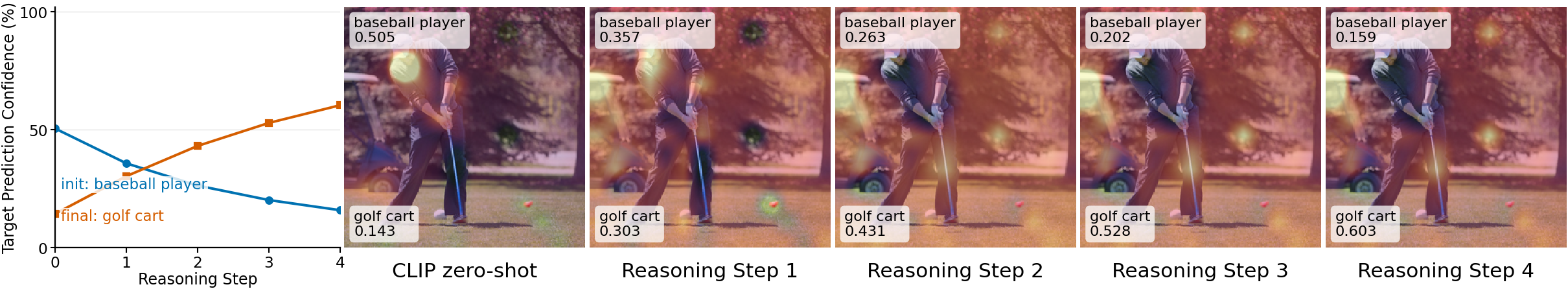}
\caption{\textbf{Wrong-to-correct transition under reasoning.} Confidence trajectories (left) and contribution maps (right) show evidence reallocating from the initial class to semantically relevant regions of the corrected class, reshaping both predictions and visual support.}
\label{fig:contribution_maps_example}
    \label{fig:image_grid}
\end{figure}

Finally, Fig.~\ref{fig:image_grid} shows an example on how visual evidence evolves during reasoning. Starting from an incorrect zero-shot prediction, PERL progressively reallocates evidence away from regions supporting the initial class and toward regions semantically consistent with the corrected prediction. Importantly, this transition is gradual rather than abrupt: intermediate reasoning steps produce coherent partial reallocations of confidence and spatial support before converging to the final prediction. This behavior suggests that iterative reasoning does not simply overwrite the classifier output, but progressively restructures the underlying visual-semantic interpretation of the image.

Extended analysis of dynamics of iterative reasoning together with additional examples are in Appendix~\ref{sec:extended_num_thoughts_analysis}.

%% file: tabs/ablation.tex
\begin{wraptable}{r}{0.41\textwidth}

\centering
\scriptsize
\setlength{\tabcolsep}{6pt}
\renewcommand{\arraystretch}{1.2}
\caption{\textbf{Ablation on PERL components.} Contribution of vision-side, text-side, and full dual-modality reasoning on base-to-novel generalization.}
\label{tab:ablation_v_t_b}
\begin{tabular}{@{}lrccc@{}}
\toprule
\textbf{Variant} & \textbf{\#Params} & \textbf{Base} & \textbf{Novel} & \textbf{HM} \\
\midrule
Vision only          & 4K          & 69.50          & 74.31          & 71.72          \\
Text only            & 3K          & 73.07          & 75.10          & 74.02          \\
\textbf{Both} (PERL) & \textbf{6K} & \textbf{73.97} & \textbf{75.74} & \textbf{74.79} \\
\bottomrule
\end{tabular}
\vspace{-0pt}
\end{wraptable}

%% file: sections/5_conclusion.tex
\label{sec:conclusion}

This paper investigated whether iterative latent reasoning can serve as a parameter-efficient adaptation mechanism for frozen contrastive vision-language models. PERL instantiates this idea through a compact, recurrent reasoning module operating within a frozen CLIP backbone. Despite using only $\sim6$K trainable parameters, PERL achieves the highest novel-class accuracy in the base-to-novel benchmark, remains competitive with substantially larger adaptation frameworks on cross-dataset transfer, and approaches methods using up to $817\times$ more parameters. 

Beyond the empirical gains, our analyses suggest that iterative reasoning performs structured, sample-specific refinement rather than a uniform distributional shift: predictions progressively stabilize, visual evidence is reallocated toward semantically consistent regions, and corrected samples exhibit substantially larger reasoning-induced drift. Overall, the results support the view that iterative latent reasoning provides an adaptation axis complementary to parameter scaling in frozen vision-language models.

\paragraph{Limitations.}
Two limitations bound our claims. First, PERL trails the strongest heavy-regime methods by $5$--$13$ points on EuroSAT and Flowers102, indicating that out-of-distribution textures and fine-grained categories still benefit from larger adaptation budgets. Second, the parameter savings come at a compute cost: PERL evaluates the refinement module $\mathcal{R}_m$ a total of $K{+}1$ times per modality (Eq.~\ref{eq:perl-recursion}), yielding $J + (K{+}1)(L{-}J)$ block evaluations versus $L$ for zero-shot CLIP. At $J{=}7$, $K{=}4$, $L{=}12$, this is ${\sim}2.67\times$ the encoder forward pass per modality. Whether this trade-off is favorable depends on whether the deployment bottleneck is storage or wall-clock latency.

\paragraph{Broader impact.}
Compact, recurrent adaptation reduces the storage and trainable-parameter burden required to specialize a single backbone across many users, domains, or deployment environments, broadening access to strong vision-language models in resource-constrained settings such as on-device personalization. This benefit trades off against higher per-example inference compute, as noted above.

%% file: sections/appendix_A_implementation_details.tex
We use OpenAI CLIP with a ViT-B/16 visual backbone (vision hidden $d_v\!=\!768$, text hidden $d_t\!=\!512$, $L\!=\!12$ Transformer blocks per encoder). The encoder split depth is $J\!=\!7$ for both modalities. Each modality uses its own thought projector $\varphi_m$ with a rank-$r$
bottleneck preceded by Layer Normalization; we use $r\!=\!1$ throughout and
$\sigma\!=\!\mathrm{GELU}$. The linear layers include biases. Per modality,
the two linear layers contribute $2d_m+d_m+1$ parameters
($W_{m,1}\!\in\!\mathbb{R}^{1\times d_m}$, $b_{m,1}\!\in\!\mathbb{R}$,
$W_{m,2}\!\in\!\mathbb{R}^{d_m\times 1}$, $b_{m,2}\!\in\!\mathbb{R}^{d_m}$),
and Layer Normalization contributes $2d_m$ affine parameters, for
$5d_m+1$ trainable parameters. Summed across modalities this gives
$5(d_v+d_t)+2=6{,}402$, which we report as the ``${\sim}6$K'' tier in the
parameter-vs-accuracy plots. All other CLIP parameters are frozen. All experiments are computed as the mean over three runs.

\paragraph{Optimization.} PERL is optimized with AdamW at learning rate $10^{-4}$, batch size $4$, and one epoch per dataset in the 16-shot setting. The anchor coefficient is set to $\lambda\!=\!1$. We perform $K\!=\!4$ refinement steps for both base-to-novel and cross-dataset evaluations.

%% file: sections/appendix_B_b2n_extended.tex
Table~\ref{tab:base2new_extended} reports the per-dataset breakdown of the
base-to-novel generalization results summarized in
Table~\ref{tab:base2new_1} of the main paper. The per-dataset view exposes
three patterns that the dataset-averaged numbers obscure.

\textbf{Novel-class robustness drives PERL's advantage.}
PERL's headline harmonic-mean number is sustained by a high
\emph{novel-class} accuracy rather than peak base-class accuracy. The average novel-class accuracy of $75.74$ is the highest of the entire table, edging out MMA ($75.73$) and exceeding MMRL ($74.56$),
MMRL$++$ ($74.82$), and MaPLe ($72.30$) by $1$ to $3.5$ points. The
decomposition makes this concrete: PERL achieves the highest novel
accuracy in the table on DTD ($62.12$) and EuroSAT ($72.25$), and is
within one point of the best on Caltech101, OxfordPets, StanfordCars,
Flowers102, SUN397, and UCF101.

\textbf{Robustness under domain shift.}
PERL is unusually strong on EuroSAT relative to its parameter budget.
CoOp ($52.03$ HM) and ProVP ($57.16$) collapse on this satellite-imagery
dataset, while PERL retains $69.62$ HM, exceeding MMRL ($66.66$) and
matching MMRL$++$ ($69.10$) with over $130\times$ fewer parameters. The
iterative reasoning loop is well-suited to domains whose visual statistics
differ markedly from natural-image pre-training: the cosine anchor
preserves CLIP's broad priors, and the loop performs a few task-specific
refinement passes on top of them.

\begin{table}[htbp]

\definecolor{highlightCol}{HTML}{FEF9E7}

\caption{\textbf{Base-to-novel generalization.} Models are trained on base classes and evaluated on both base and novel classes; HM is the harmonic mean. \textbf{\#Params} reports trainable parameters; \textbf{$\times$PERL} the ratio to PERL's parameter count; \textbf{$\Delta$HM} the average-HM gap vs.\ PERL (positive = method exceeds PERL; negative = PERL exceeds method). 
}
\label{tab:base2new_extended}
\centering
\scriptsize
\setlength{\tabcolsep}{2pt}
\renewcommand{\arraystretch}{1.4} 

\setlength{\aboverulesep}{0pt}
\setlength{\belowrulesep}{0pt}
\resizebox{\textwidth}{!}{
\begin{tabular}{@{}l r >{\columncolor{highlightCol}}r ccc >{\columncolor{highlightCol}}c ccc ccc ccc@{}}
\toprule
\textbf{Method} & \textbf{\#Params} & \textbf{$\times$PERL}
& \multicolumn{4}{c}{\textbf{Average}}
& \multicolumn{3}{c}{\textbf{ImageNet}}
& \multicolumn{3}{c}{\textbf{Caltech101}}
& \multicolumn{3}{c}{\textbf{OxfordPets}} \\
\cmidrule(lr){4-7}\cmidrule(lr){8-10}\cmidrule(lr){11-13}\cmidrule(lr){14-16}
 & & & Base & Novel & HM & $\Delta$HM & Base & Novel & HM & Base & Novel & HM & Base & Novel & HM \\
\midrule

CLIP$_{\text{ICML'21}}$ & 0 & --- & 69.34 & 74.35 & 71.64 & $-3.15$ & 72.43 & 68.14 & 70.22 & 96.84 & 94.00 & 95.40 & 91.17 & 97.26 & 94.12 \\
\midrule

MMRL$_{\text{CVPR'25}}$    & 4.9M & 817$\times$ & 70.92 & 74.27 & 72.45 & $-2.34$ & 76.47 & 71.33 & 73.81 & 96.90 & 92.73 & 94.77 & 88.83 & 95.53 & 92.06 \\
MMRL++$_{\text{IJCV'25}}$  & 813K & 136$\times$ & 70.92 & 74.27 & 72.45 & $-1.72$ & 76.47 & 71.33 & 73.81 & 97.37 & 94.00 & 95.66 & 91.10 & 94.63 & 92.83 \\
MaPLe$_{\text{CVPR'23}}$  & 3.5M & 583$\times$ & 75.69 & 72.30 & 73.79 & $-1.00$ & 75.80 & 71.27 & 73.47 & 95.60 & 93.80 & 94.69 & 94.90 & 97.73 & 96.29 \\
MMA$_{\text{CVPR'24}}$    & 675K & 113$\times$        & 75.18 & 75.73 & 75.23 & $+0.44$ & 76.10 & 71.26 & 73.60 & 97.30 & 94.13 & 95.69 & 92.96 & 96.96 & 94.92 \\
TCP$_{\text{CVPR'24}}$    & 331K & 55$\times$ & 64.85 & 69.04 & 66.80 & $-7.99$ & 69.63 & 64.10 & 66.75 & 87.23 & 91.73 & 89.42 & 86.70 & 90.43 & 88.53 \\
ProVP$_{\text{IJCV'24}}$  & 147K & 25$\times$ & 72.00 & 73.65 & 72.76 & $-2.03$ & 74.19 & 69.28 & 71.65 & 96.92 & 93.41 & 95.13 & 91.99 & 96.81 & 94.34 \\

CoCoOp$_{\text{CVPR'22}}$    & 35K & 6$\times$ & 74.90 & 72.35 & 73.43 & $-1.36$ & 75.39 & 70.70 & 72.97 & 97.10 & 93.63 & 95.33 & 94.60 & 97.26 & 95.91 \\
PromptSRC$_{\text{ICCV'23}}$ & 46K & 8$\times$       & 77.45 & 74.34 & 75.66 & $+0.87$ & 75.60 & 70.93 & 73.19 & 97.83 & 94.76 & 96.27 & 95.13 & 97.30 & 96.20 \\

KgCoOp$_{\text{CVPR'23}}$ & 2K & 0.33$\times$ & 66.36 & 74.22 & 69.87 & $-4.92$ & 50.20 & 69.83 & 58.41 & 97.27 & 94.14 & 95.68 & 89.40 & 96.85 & 92.98 \\
CoOp$_{\text{IJCV'22}}$   & 2K & 0.33$\times$ & 70.73 & 70.74 & 70.67 & $-4.12$ & 75.30 & 70.57 & 72.86 & 96.93 & 93.60 & 95.24 & 90.13 & 96.67 & 93.29 \\
\midrule
\textbf{PERL (Ours)}      & 6K & 1$\times$    & 73.97 & 75.74 & 74.79 & ---     & 75.16 & 70.87 & 72.95 & 97.91 & 94.58 & 96.22 & 94.77 & 97.48 & 96.11 \\
\bottomrule
\end{tabular}}
\vspace{4pt}\\
{\fontsize{7pt}{8pt}\selectfont
   
\begin{tabular*}{\textwidth}{@{\extracolsep{\fill}} l ccc ccc ccc ccc @{}}
\toprule
\textbf{Method} 
& \multicolumn{3}{c}{\textbf{StanfordCars}}
& \multicolumn{3}{c}{\textbf{Flowers102}}
& \multicolumn{3}{c}{\textbf{Food101}}
& \multicolumn{3}{c}{\textbf{FGVCAircraft}} \\
\cmidrule(lr){2-4}\cmidrule(lr){5-7}\cmidrule(lr){8-10}\cmidrule(lr){11-13}
 & Base & Novel & HM & Base & Novel & HM & Base & Novel & HM & Base & Novel & HM \\
\midrule
CLIP$_{\text{ICML'21}}$ & 63.37 & 74.89 & 68.65 & 72.08 & 77.80 & 74.83 & 90.10 & 91.22 & 90.66 & 27.19 & 36.29 & 31.09 \\
\midrule

MMRL$_{\text{CVPR'25}}$    & 63.67 & 74.97 & 68.86 & 70.23 & 76.93 & 73.43 & 89.07 & 90.93 & 89.99 & 28.93 & 35.77 & 31.99 \\
MMRL++$_{\text{IJCV'25}}$  & 65.33 & 74.80 & 69.75 & 72.00 & 77.70 & 74.74 & 89.03 & 90.37 & 89.69 & 29.97 & 34.83 & 32.22 \\
MaPLe$_{\text{CVPR'23}}$   & 67.50 & 74.00 & 70.60 & 86.47 & 75.00 & 80.33 & 90.50 & 91.80 & 91.15 & 17.57 & 11.90 & 14.19 \\
MMA$_{\text{CVPR'24}}$     & 69.37 & 75.20 & 72.17 & 90.06 & 77.10 & 83.08 & 90.53 & 91.60 & 91.06 & 33.87 & 35.50 & 34.67 \\
TCP$_{\text{CVPR'24}}$     & 60.03 & 71.83 & 65.40 & 65.43 & 70.83 & 68.02 & 88.66 & 89.63 & 89.14 & 18.53 & 23.63 & 20.77 \\
ProVP$_{\text{IJCV'24}}$   & 66.08 & 74.97 & 70.24 & 75.08 & 76.24 & 75.66 & 90.07 & 91.12 & 90.59 & 30.01 & 35.03 & 32.33 \\

CoCoOp$_{\text{CVPR'22}}$    & 66.46 & 75.00 & 70.47 & 82.90 & 74.39 & 78.41 & 90.23 & 90.70 & 90.46 & 23.60 & 24.70 & 24.14 \\
PromptSRC$_{\text{ICCV'23}}$ & 67.43 & 75.80 & 71.37 & 88.63 & 76.30 & 82.00 & 90.36 & 91.73 & 91.04 & 32.23 & 35.03 & 33.57 \\

KgCoOp$_{\text{CVPR'23}}$ & 63.78 & 74.92 & 68.90 & 69.27 & 76.86 & 72.87 & 89.44 & 90.70 & 90.07 & 27.63 & 33.09 & 30.11 \\
CoOp$_{\text{IJCV'22}}$   & 66.06 & 74.10 & 69.85 & 77.13 & 74.10 & 75.58 & 89.27 & 90.50 & 89.88 & 20.90 & 20.09 & 20.49 \\
\midrule
\textbf{PERL (Ours)} & 66.64 & 74.57 & 70.38 & 78.35 & 76.85 & 77.59 & 90.60 & 91.67 & 91.13 & 31.29 & 35.09 & 33.08 \\
\bottomrule
\end{tabular*}}
\vspace{4pt}\\

{\fontsize{7pt}{8pt}\selectfont
\begin{tabular*}{\textwidth}{@{\extracolsep{\fill}} l ccc ccc ccc ccc @{}}
\toprule
\textbf{Method} 
& \multicolumn{3}{c}{\textbf{SUN397}}
& \multicolumn{3}{c}{\textbf{DTD}}
& \multicolumn{3}{c}{\textbf{EuroSAT}}
& \multicolumn{3}{c}{\textbf{UCF101}} \\
\cmidrule(lr){2-4}\cmidrule(lr){5-7}\cmidrule(lr){8-10}\cmidrule(lr){11-13}
 & Base & Novel & HM & Base & Novel & HM & Base & Novel & HM & Base & Novel & HM \\
\midrule
CLIP$_{\text{ICML'21}}$ & 69.36 & 75.35 & 72.23 & 53.24 & 61.30 & 56.99 & 56.48 & 64.05 & 60.03 & 70.53 & 77.50 & 73.85 \\
\midrule

MMRL$_{\text{CVPR'25}}$    & 72.50 & 77.80 & 75.06 & 54.03 & 58.17 & 56.02 & 68.60 & 64.83 & 66.66 & 70.93 & 77.93 & 74.27 \\
MMRL++$_{\text{IJCV'25}}$  & 72.90 & 77.40 & 75.08 & 55.43 & 56.23 & 55.83 & 67.73 & 70.53 & 69.10 & 72.57 & 77.70 & 75.05 \\
MaPLe$_{\text{CVPR'23}}$   & 77.40 & 78.37 & 77.88 & 68.37 & 60.27 & 64.06 & 80.30 & 64.80 & 71.72 & 78.23 & 76.33 & 77.27 \\
MMA$_{\text{CVPR'24}}$     & 79.06 & 78.46 & 78.76 & 67.63 & 64.30 & 65.92 & 50.16 & 69.93 & 58.42 & 79.90 & 78.63 & 79.26 \\
TCP$_{\text{CVPR'24}}$     & 62.66 & 66.73 & 64.63 & 44.73 & 47.20 & 45.93 & 60.60 & 69.06 & 64.55 & 69.10 & 74.30 & 71.61 \\
ProVP$_{\text{IJCV'24}}$   & 73.32 & 76.61 & 74.93 & 62.46 & 61.47 & 61.96 & 55.74 & 58.65 & 57.16 & 76.11 & 76.56 & 76.33 \\

CoCoOp$_{\text{CVPR'22}}$    & 75.70 & 77.63 & 76.65 & 66.07 & 59.30 & 62.50 & 75.13 & 56.07 & 64.22 & 76.77 & 76.50 & 76.63 \\
PromptSRC$_{\text{ICCV'23}}$ & 78.33 & 79.23 & 78.78 & 66.66 & 52.76 & 58.90 & 80.63 & 64.33 & 71.56 & 79.17 & 79.57 & 79.37 \\

KgCoOp$_{\text{CVPR'23}}$ & 69.90 & 75.89 & 72.77 & 54.71 & 59.18 & 56.86 & 50.20 & 69.83 & 58.41 & 68.17 & 75.14 & 71.49 \\
CoOp$_{\text{IJCV'22}}$   & 76.03 & 77.30 & 76.66 & 60.13 & 53.53 & 56.64 & 50.43 & 53.73 & 52.03 & 75.73 & 74.00 & 74.86 \\
\midrule
\textbf{PERL (Ours)} & 75.74 & 78.24 & 76.97 & 59.69 & 62.12 & 60.87 & 67.25 & 72.25 & 69.62 & 76.23 & 79.39 & 77.78 \\
\bottomrule
\end{tabular*}}
\end{table}

%% file: sections/appendix_C_technical_supplementary.tex
PERL shares the high-level intuition of weight-shared iterative computation with Universal Transformers~\cite{universaltransformer}, deep equilibrium models~\cite{deq}, and looped transformers~\cite{loopedtransformers,loopedtransformersreasoning}, but differs from them on four concrete axes. First, recurrence in PERL acts on a single pooled CLS/EOS \emph{vector} passed through a tiny MLP, not on the full token sequence through a Transformer block as in Universal Transformers and looped transformers~\cite{loopedtransformers,loopedtransformersreasoning}. Second, the recurrent projector updates a \emph{frozen} refinement stack rather than being co-trained with it, so additional refinement steps add no parameters. Third, PERL targets contrastive/discriminative readout rather than autoregressive generation, distinguishing it from Coconut~\cite{coconut} and Quiet-STaR~\cite{quietstar}. Fourth, the zero-shot anchor introduced above is specific to grounding refinement in CLIP's pretrained geometry and has no counterpart in looped or equilibrium formulations. Relative to MMRL/MMRL++~\cite{mmrl,mmrl++}, the closest CLIP-adaptation neighbors, PERL replaces a fixed bank of trainable representation tokens with thought tokens \emph{generated by} a parameter-shared $\varphi_m$ from the current pooled state, so adaptation capacity grows with the number of recurrent steps rather than with the size of the token bank.

%% file: sections/appendix_D_original_baselines.tex
The main results in Tables~\ref{tab:base2new_1} and~\ref{tab:cross_dataset_1}
report every method under a unified protocol --- single-epoch training, the
same 16-shot supervision per class --- to isolate the architectural choice from the
epoch budget. To check that this protocol is not unfairly
disadvantageous to the prompt-tuning and adapter baselines, we report each competitor's results in their default training epochs in Tables~\ref{tab:base2new_default_competitors} and
\ref{tab:cross_dataset_default_competitors}. PERL is left unchanged at its single-epoch, $6$K-parameter configuration.

The overview matches the main paper. On base-to-novel
generalization, PERL's novel-class accuracy remains among the highest in
the table even when larger-capacity baselines are granted their full training
budget, and its average HM is competitive with all but the two strongest
methods (PromptSRC and MMA). The parameter--efficiency advantage is in
fact sharper than in the main protocol: methods that surpass PERL by less
than $1.5$ HM points use $8\times$ to $817\times$ more trainable
parameters \emph{and} two orders of magnitude more gradient updates over
the target dataset. On cross-dataset transfer, the relative ordering
between methods is preserved, with PERL exceeding CoOp, CoCoOp, KgCoOp,
MMA, ProVP, and TCP on the target average, and trailing MMRL and MMRL++
by a margin comparable to that of the main results. Taken together, these
tables confirm that PERL's standing on the
parameter--accuracy--training-time trade-off is not an artifact of the
unified evaluation protocol but a property of the architecture itself: the
baselines retain their reported strengths under the original recipes, yet
PERL --- trained for a single epoch with a fraction of the parameters ---
remains competitive on every benchmark.

\input{tabs/base2new_default_competitors}
\input{tabs/cross_dataset_default_competitors}

%% file: tabs/base2new_default_competitors.tex
\begin{table}[htbp]
\definecolor{highlightCol}{HTML}{FEF9E7}
\caption{\textbf{Base-to-novel generalization.} Models are trained on base classes and evaluated on both base and novel classes; HM is the harmonic mean. \textbf{\#Params} reports trainable parameters; \textbf{$\times$PERL} the ratio to PERL's parameter count; \textbf{$\Delta$HM} the average-HM gap vs.\ PERL (positive = method exceeds PERL; negative = PERL exceeds method). All baseline numbers are reproduced under their default setting.}
\label{tab:base2new_default_competitors}
\centering
\scriptsize
\setlength{\tabcolsep}{2pt}
\renewcommand{\arraystretch}{1.4}
\setlength{\aboverulesep}{0pt}
\setlength{\belowrulesep}{0pt}
\resizebox{\textwidth}{!}{
\begin{tabular}{@{}l r >{\columncolor{highlightCol}}r ccc >{\columncolor{highlightCol}}c ccc ccc ccc@{}}
\toprule
\textbf{Method} & \textbf{\#Params} & \textbf{$\times$PERL}
& \multicolumn{4}{c}{\textbf{Average}}
& \multicolumn{3}{c}{\textbf{ImageNet}}
& \multicolumn{3}{c}{\textbf{Caltech101}}
& \multicolumn{3}{c}{\textbf{OxfordPets}} \\
\cmidrule(lr){4-7}\cmidrule(lr){8-10}\cmidrule(lr){11-13}\cmidrule(lr){14-16}
 & & & Base & Novel & HM & $\Delta$HM & Base & Novel & HM & Base & Novel & HM & Base & Novel & HM \\
\midrule
CLIP$_{\text{ICML'21}}$ & 0 & --- & 69.34 & 74.22 & 71.70 & $-3.09$ & 72.43 & 68.14 & 70.22 & 96.84 & 94.00 & 95.40 & 91.17 & 97.26 & 94.12 \\
\midrule
MMRL$_{\text{CVPR'25}}$ & 4.9M & 817$\times$ & 85.68 & 77.16 & 81.20 & $+6.41$ & 77.90 & 71.30 & 74.45 & 98.97 & 94.50 & 96.68 & 95.90 & 97.60 & 96.74 \\
MMRL++$_{\text{IJCV'25}}$ & 813K & 136$\times$ & 85.53 & 78.32 & 81.77 & $+6.98$ & 77.63 & 71.50 & 74.44 & 99.07 & 94.53 & 96.75 & 95.60 & 97.43 & 96.51 \\
MaPLe$_{\text{CVPR'23}}$ & 3.5M & 583$\times$ & 82.28 & 75.14 & 78.55 & $+3.76$ & 76.66 & 70.54 & 73.47 & 97.74 & 94.36 & 96.02 & 95.43 & 97.76 & 96.58 \\
MMA$_{\text{CVPR'24}}$ & 675K & 113$\times$ & 83.20 & 76.80 & 79.87 & $+5.08$ & 77.31 & 71.00 & 74.02 & 98.40 & 94.00 & 96.15 & 95.40 & 98.07 & 96.72 \\
TCP$_{\text{CVPR'24}}$ & 331K & 55$\times$ & 84.13 & 75.36 & 79.51 & $+4.72$ & 77.27 & 69.87 & 73.38 & 98.23 & 94.67 & 96.42 & 94.67 & 97.20 & 95.92 \\
ProVP$_{\text{IJCV'24}}$ & 147K & 25$\times$ & 85.20 & 73.22 & 78.76 & $+3.97$ & 75.82 & 69.21 & 72.36 & 98.92 & 94.21 & 96.51 & 95.87 & 97.65 & 96.75 \\
CoCoOp$_{\text{CVPR'22}}$ & 35K & 6$\times$ & 80.47 & 71.69 & 75.83 & $+1.04$ & 75.98 & 70.43 & 73.10 & 97.96 & 93.81 & 95.84 & 95.20 & 97.69 & 96.43 \\
PromptSRC$_{\text{ICCV'23}}$ & 46K & 8$\times$ & 84.26 & 76.10 & 79.97 & $+5.18$ & 77.60 & 70.73 & 74.01 & 98.10 & 94.03 & 96.02 & 95.33 & 97.30 & 96.30 \\
KgCoOp$_{\text{CVPR'23}}$ & 2K & 0.33$\times$ & 80.73 & 73.60 & 77.00 & $+2.21$ & 75.83 & 69.96 & 72.78 & 97.72 & 94.39 & 96.03 & 94.65 & 97.76 & 96.18 \\
CoOp$_{\text{IJCV'22}}$ & 2K & 0.33$\times$ & 82.69 & 63.22 & 71.66 & $-3.13$ & 76.47 & 67.88 & 71.92 & 98.00 & 89.81 & 93.73 & 93.67 & 95.29 & 94.47 \\
\midrule
\textbf{PERL (Ours)} & 6K & 1$\times$ & 73.97 & 75.74 & 74.79 & --- & 75.16 & 70.87 & 72.95 & 97.91 & 94.58 & 96.22 & 94.76 & 97.48 & 96.11 \\
\bottomrule
\end{tabular}}
\vspace{4pt}\\
{\fontsize{7pt}{8pt}\selectfont
\begin{tabular*}{\textwidth}{@{\extracolsep{\fill}} l ccc ccc ccc ccc @{}}
\toprule
\textbf{Method}
& \multicolumn{3}{c}{\textbf{StanfordCars}}
& \multicolumn{3}{c}{\textbf{Flowers102}}
& \multicolumn{3}{c}{\textbf{Food101}}
& \multicolumn{3}{c}{\textbf{FGVCAircraft}} \\
\cmidrule(lr){2-4}\cmidrule(lr){5-7}\cmidrule(lr){8-10}\cmidrule(lr){11-13}
 & Base & Novel & HM & Base & Novel & HM & Base & Novel & HM & Base & Novel & HM \\
\midrule
CLIP$_{\text{ICML'21}}$ & 63.37 & 74.89 & 68.65 & 72.08 & 77.80 & 74.83 & 90.10 & 91.22 & 90.66 & 27.19 & 36.29 & 31.09 \\
\midrule
MMRL$_{\text{CVPR'25}}$ & 81.30 & 75.07 & 78.06 & 98.97 & 77.27 & 86.78 & 90.57 & 91.50 & 91.03 & 46.30 & 37.03 & 41.15 \\
MMRL++$_{\text{IJCV'25}}$ & 81.33 & 75.27 & 78.18 & 98.53 & 77.90 & 87.01 & 90.47 & 91.73 & 91.10 & 46.40 & 38.77 & 42.24 \\
MaPLe$_{\text{CVPR'23}}$ & 72.94 & 74.00 & 73.47 & 95.92 & 72.46 & 82.56 & 90.71 & 92.05 & 91.38 & 37.44 & 35.61 & 36.50 \\
MMA$_{\text{CVPR'24}}$ & 78.50 & 73.10 & 75.70 & 97.77 & 75.93 & 85.48 & 90.13 & 91.30 & 90.71 & 40.57 & 36.33 & 38.33 \\
TCP$_{\text{CVPR'24}}$ & 80.80 & 74.13 & 77.32 & 97.73 & 75.57 & 85.23 & 90.57 & 91.37 & 90.97 & 41.97 & 34.43 & 37.83 \\
ProVP$_{\text{IJCV'24}}$ & 80.43 & 67.96 & 73.67 & 98.42 & 72.06 & 83.20 & 90.32 & 90.91 & 90.61 & 47.08 & 29.87 & 36.55 \\
CoCoOp$_{\text{CVPR'22}}$ & 70.49 & 73.59 & 72.01 & 94.87 & 71.75 & 81.71 & 90.70 & 91.29 & 90.99 & 33.41 & 23.71 & 27.74 \\
PromptSRC$_{\text{ICCV'23}}$ & 78.27 & 74.97 & 76.58 & 98.07 & 76.50 & 85.95 & 90.67 & 91.53 & 91.10 & 42.73 & 37.87 & 40.15 \\
KgCoOp$_{\text{CVPR'23}}$ & 71.76 & 75.04 & 73.36 & 95.00 & 74.73 & 83.65 & 90.50 & 91.70 & 91.10 & 36.21 & 33.55 & 34.83 \\
CoOp$_{\text{IJCV'22}}$ & 78.12 & 60.40 & 68.13 & 97.60 & 59.67 & 74.06 & 88.33 & 82.26 & 85.19 & 40.44 & 22.30 & 28.75 \\
\midrule
\textbf{PERL (Ours)} & 66.64 & 74.57 & 70.38 & 78.35 & 76.85 & 77.59 & 90.60 & 91.67 & 91.13 & 31.29 & 35.09 & 33.08 \\
\bottomrule
\end{tabular*}}
\vspace{4pt}\\
{\fontsize{7pt}{8pt}\selectfont
\begin{tabular*}{\textwidth}{@{\extracolsep{\fill}} l ccc ccc ccc ccc @{}}
\toprule
\textbf{Method}
& \multicolumn{3}{c}{\textbf{SUN397}}
& \multicolumn{3}{c}{\textbf{DTD}}
& \multicolumn{3}{c}{\textbf{EuroSAT}}
& \multicolumn{3}{c}{\textbf{UCF101}} \\
\cmidrule(lr){2-4}\cmidrule(lr){5-7}\cmidrule(lr){8-10}\cmidrule(lr){11-13}
 & Base & Novel & HM & Base & Novel & HM & Base & Novel & HM & Base & Novel & HM \\
\midrule
CLIP$_{\text{ICML'21}}$ & 69.36 & 75.35 & 72.23 & 53.24 & 61.30 & 56.99 & 56.48 & 64.05 & 60.03 & 70.53 & 77.50 & 73.85 \\
\midrule
MMRL$_{\text{CVPR'25}}$ & 83.20 & 79.30 & 81.20 & 85.67 & 65.00 & 73.92 & 95.60 & 80.17 & 87.21 & 88.10 & 80.07 & 83.89 \\
MMRL++$_{\text{IJCV'25}}$ & 83.03 & 79.60 & 81.28 & 85.47 & 65.97 & 74.46 & 95.93 & 88.27 & 91.94 & 87.37 & 80.53 & 83.81 \\
MaPLe$_{\text{CVPR'23}}$ & 80.82 & 78.70 & 79.75 & 80.36 & 59.18 & 68.16 & 94.07 & 73.23 & 82.35 & 83.00 & 78.66 & 80.77 \\
MMA$_{\text{CVPR'24}}$ & 82.27 & 78.57 & 80.38 & 83.20 & 65.63 & 73.38 & 85.46 & 82.34 & 83.87 & 86.23 & 80.03 & 83.01 \\
TCP$_{\text{CVPR'24}}$ & 82.63 & 78.20 & 80.35 & 82.77 & 58.07 & 68.25 & 91.63 & 74.73 & 82.32 & 87.13 & 80.77 & 83.83 \\
ProVP$_{\text{IJCV'24}}$ & 80.67 & 76.11 & 78.32 & 83.95 & 59.06 & 69.34 & 97.12 & 72.91 & 83.29 & 88.56 & 75.55 & 81.54 \\
CoCoOp$_{\text{CVPR'22}}$ & 79.74 & 76.86 & 78.27 & 77.01 & 56.00 & 64.85 & 87.49 & 60.04 & 71.21 & 82.33 & 73.45 & 77.64 \\
PromptSRC$_{\text{ICCV'23}}$ & 82.67 & 78.47 & 80.52 & 83.37 & 62.97 & 71.75 & 92.90 & 73.90 & 82.32 & 87.10 & 78.80 & 82.74 \\
KgCoOp$_{\text{CVPR'23}}$ & 80.29 & 76.53 & 78.36 & 77.55 & 54.99 & 64.35 & 85.64 & 64.34 & 73.48 & 82.89 & 76.67 & 79.66 \\
CoOp$_{\text{IJCV'22}}$ & 80.60 & 65.89 & 72.51 & 79.44 & 41.18 & 54.24 & 92.19 & 54.74 & 68.69 & 84.69 & 56.05 & 67.46 \\
\midrule
\textbf{PERL (Ours)} & 75.74 & 78.24 & 76.97 & 59.69 & 62.12 & 60.87 & 67.25 & 72.25 & 69.62 & 76.23 & 79.39 & 77.78 \\
\bottomrule
\end{tabular*}}
\end{table}

%% file: tabs/cross_dataset_default_competitors.tex
\begin{table}[htbp]
\definecolor{highlightCol}{HTML}{FEF9E7}   
\caption{\textbf{Cross-dataset evaluation.} Models are trained on ImageNet (source) and evaluated zero-shot on 10 target datasets. \textbf{\#Params} reports trainable parameters; \textbf{$\times$PERL} the ratio to PERL's count; \textbf{$\Delta$Avg.} the gap in mean target accuracy vs.\ PERL (positive = method exceeds PERL; negative = PERL exceeds method). All baseline numbers are reproduced under their default setting.}
\label{tab:cross_dataset_default_competitors}
\centering
\scriptsize
\setlength{\tabcolsep}{3pt}
\renewcommand{\arraystretch}{1.4}
\setlength{\aboverulesep}{0pt}
\setlength{\belowrulesep}{0pt}
\resizebox{\textwidth}{!}{%
\begin{tabular}{@{}l r >{\columncolor{highlightCol}}r c cccccccccc c >{\columncolor{highlightCol}}r@{}}
\toprule
\textbf{Method} & \textbf{\#Params} & \textbf{$\times$PERL} & \textbf{Source} & \multicolumn{10}{c}{\textbf{Target}} & \textbf{Average} & \textbf{$\Delta$Avg.} \\
\cmidrule(lr){4-4}\cmidrule(lr){5-14}
 & & &
\rotatebox{90}{ImageNet} &
\rotatebox{90}{Caltech101} &
\rotatebox{90}{OxfordPets} &
\rotatebox{90}{StanfordCars} &
\rotatebox{90}{Flowers102} &
\rotatebox{90}{Food101} &
\rotatebox{90}{FGVCAircraft} &
\rotatebox{90}{SUN397} &
\rotatebox{90}{DTD} &
\rotatebox{90}{EuroSAT} &
\rotatebox{90}{UCF101} & & \\
\midrule
MMRL$_{\text{CVPR'25}}$ & 4.9M & 817$\times$ & 72.03 & 94.67 & 91.43 & 66.10 & 72.77 & 86.40 & 26.30 & 67.57 & 45.90 & 53.10 & 68.27 & 67.25 & $+1.69$ \\
MMRL++$_{\text{IJCV'25}}$ & 813K & 136$\times$ & 71.87 & 94.63 & 91.43 & 66.60 & 73.53 & 86.73 & 26.07 & 67.77 & 46.13 & 53.00 & 69.03 & 67.49 & $+1.93$ \\
MaPLe$_{\text{CVPR'23}}$ & 3.5M & 583$\times$ & 70.72 & 93.53 & 90.49 & 65.57 & 72.23 & 86.20 & 24.74 & 67.01 & 46.49 & 48.06 & 68.69 & 66.30 & $+0.74$ \\
MMA$_{\text{CVPR'24}}$ & 675K & 113$\times$ & 71.00 & 93.80 & 90.30 & 66.13 & 72.07 & 86.12 & 25.33 & 68.17 & 46.57 & 49.24 & 68.32 & 66.61 & $+1.05$ \\
TCP$_{\text{CVPR'24}}$ & 331K & 55$\times$ & 71.40 & 93.97 & 91.25 & 64.69 & 71.21 & 86.69 & 23.45 & 67.15 & 44.35 & 51.45 & 68.73 & 66.29 & $+0.73$ \\
CoCoOp$_{\text{CVPR'22}}$ & 35K & 6$\times$ & 71.02 & 94.43 & 90.14 & 65.32 & 71.88 & 86.06 & 22.94 & 67.36 & 45.73 & 45.37 & 68.21 & 65.74 & $+0.18$ \\
PromptSRC$_{\text{ICCV'23}}$ & 46K & 8$\times$ & 71.27 & 93.60 & 90.25 & 65.70 & 70.25 & 86.15 & 23.90 & 67.10 & 46.87 & 45.50 & 68.75 & 65.81 & $+0.25$ \\
KgCoOp$_{\text{CVPR'23}}$ & 2K & 0.33$\times$ & 70.66 & 93.70 & 89.83 & 65.41 & 70.01 & 86.36 & 22.51 & 66.16 & 46.35 & 46.04 & 68.50 & 65.96 & $+0.40$ \\
CoOp$_{\text{IJCV'22}}$ & 2K & 0.33$\times$ & 71.51 & 93.70 & 89.14 & 64.51 & 68.71 & 85.30 & 18.47 & 64.15 & 41.92 & 46.39 & 66.55 & 63.88 & $-1.68$ \\
\midrule
\textbf{PERL (Ours)} & 6K & 1$\times$ & 70.25 & 94.33 & 90.36 & 66.92 & 70.79 & 86.37 & 24.60 & 67.46 & 45.72 & 40.01 & 69.05 & 65.56 & --- \\
\bottomrule
\end{tabular}}
\end{table}

%% file: sections/appendix_E_ood.tex
Table~\ref{tab:ood} reports robustness to natural and synthetic
distribution shifts on the ImageNet label space. Models are trained on
ImageNet and evaluated zero-shot on four standard variants: ImageNet-V2
(natural-image test set), ImageNet-Sketch (sketches of all $1000$
classes), ImageNet-A ($200$-class natural adversarial images), and
ImageNet-R (renditions of $200$ classes). Baselines are
reported in their original paper setting.

PERL reaches an average target accuracy of $59.95$, within $0.7$ points
of the strongest baselines (PromptSRC $60.65$, MMRL$++$ $60.60$,
MMRL $60.59$, MMA $60.48$) despite using $8\times$ to $817\times$ fewer
trainable parameters and a single training epoch. It exceeds CLIP
zero-shot ($57.18$) by $2.77$ points and outperforms or matches the
early prompt-tuning baselines CoOp ($59.28$) and CoCoOp ($59.91$).
Per-variant, the gap to the best is below one point on every dataset:
ImageNet-V2 ($-0.99$), ImageNet-Sketch ($-0.80$), ImageNet-A ($-0.92$),
and ImageNet-R ($-0.72$).

Two observations are worth noting. First, PERL's source ImageNet
accuracy ($70.26$) is also $1$--$2$ points below the strongest baselines
(MMRL $72.03$, CoOp $71.51$). The Source--Target gap of $10.31$ points
is smaller than that of CoOp ($12.23$) and comparable to PromptSRC
($10.62$).

Second, the residual deficit is most concentrated on ImageNet-A, the
adversarially curated subset on which CLIP zero-shot is weakest. The
low-rank thought projector evidently lacks the capacity to disambiguate
examples specifically designed to fool natural-image classifiers --- a
weakness shared with the small prompt-tuning baselines KgCoOp ($50.69$)
and CoCoOp ($50.63$). Overall, the table confirms that the iterative
reasoning loop, anchored to CLIP's zero-shot manifold, retains the
backbone's pre-trained robustness while remaining parameter- and
training-efficient: PERL closes the bulk of the OOD gap to heavy
baselines at a fraction of the cost.
\input{tabs/ood}

%% file: tabs/ood.tex

\begin{table}[ht]
\centering
\caption{distribution shift evaluation with state-of-the art methods in their default setting.}
\label{tab:ood}
\scriptsize
\setlength{\tabcolsep}{8pt}
\renewcommand{\arraystretch}{1.4}
\begin{tabular}{lcccccc}
\toprule
\multirow{2}{*}{Method} & Source & \multicolumn{5}{c}{Target} \\
\cmidrule(lr){2-2} \cmidrule(lr){3-7}
 & ImNet & -V2 & -S & -A & -R & Average \\
\midrule
CLIP      & 66.73 & 60.83 & 46.15 & 47.77 & 73.96 & 57.18 \\
CoOp      & 71.51 & 64.20 & 47.99 & 49.71 & 75.21 & 59.28 \\
CoCoOp    & 71.02 & 64.07 & 48.75 & 50.63 & 76.18 & 59.91 \\
KgCoOp    & 71.20 & 64.10 & 48.97 & 50.69 & 76.70 & 60.12 \\
MaPLe     & 70.72 & 64.07 & 49.15 & 50.90 & 76.98 & 60.28 \\
PromptSRC & 71.27 & 64.35 & 49.55 & 50.90 & 77.80 & 60.65 \\
MMA       & 71.00 & 64.33 & 49.13 & 51.12 & 77.32 & 60.48 \\
MMRL      & 72.03 & 64.47 & 49.17 & 51.20 & 77.53 & 60.59 \\
MMRL++    & 71.87 & 64.67 & 49.30 & 51.00 & 77.43 & 60.60 \\
\textbf{PERL}      & 70.26 & 63.68 & 48.75 & 50.28 & 77.08 & 59.95 \\
\bottomrule
\end{tabular}
\end{table}

%% file: sections/appendix_F_extended_analysis.tex
\begin{figure}[htbp]
    \centering
    \includegraphics[width=0.95\textwidth]{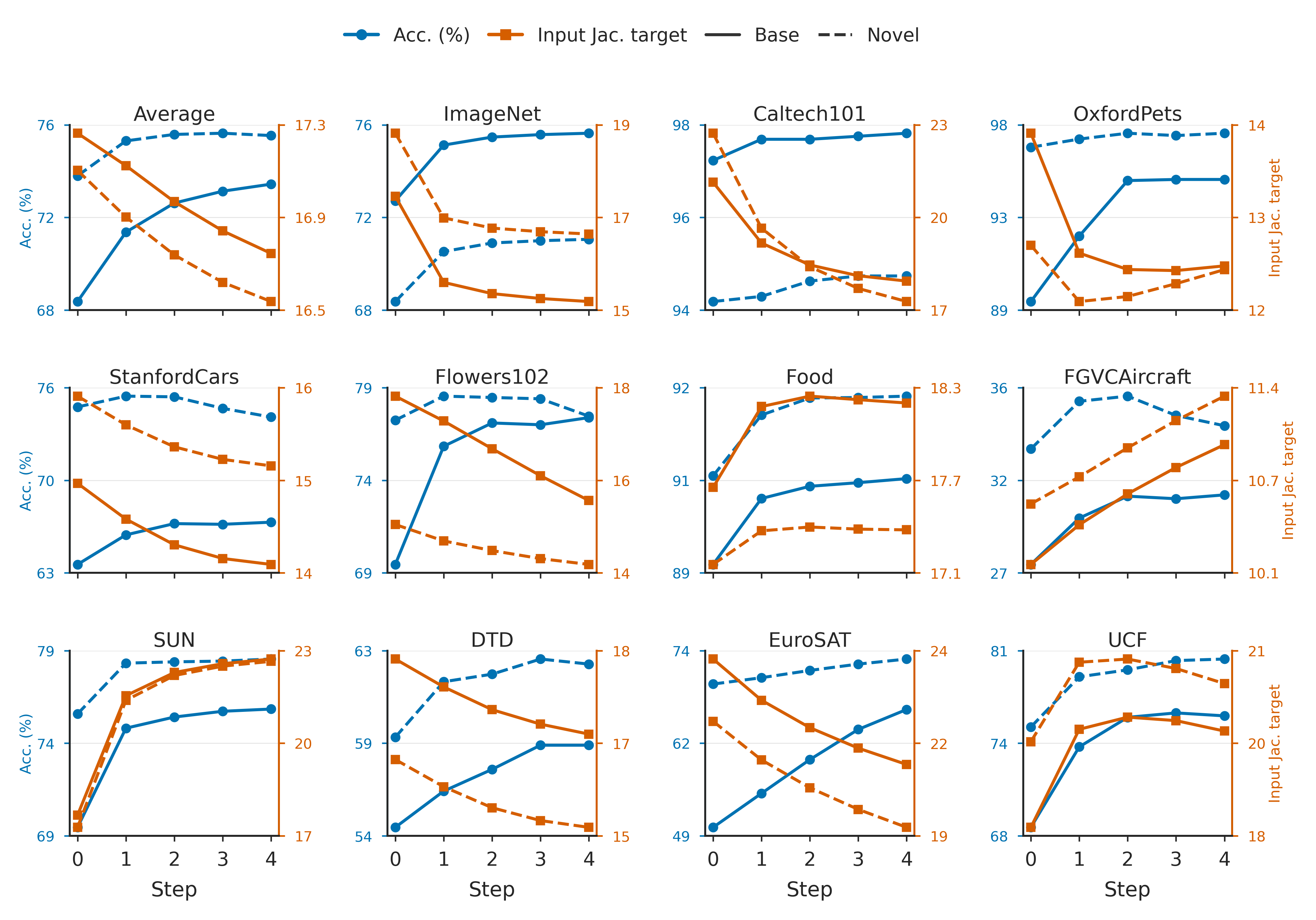}
    \caption{Accuracy and Jacobian norm of the target logit with respect to the input in the base-to-novel evaluation settings. We report each metric for every dataset and across varying numbers of reasoning steps.}
    \label{fig:accuracy-input-jacobian-target-b2n-per-dataset}
\end{figure}
\begin{figure}[htbp]
    \centering
    \includegraphics[width=0.95\textwidth]{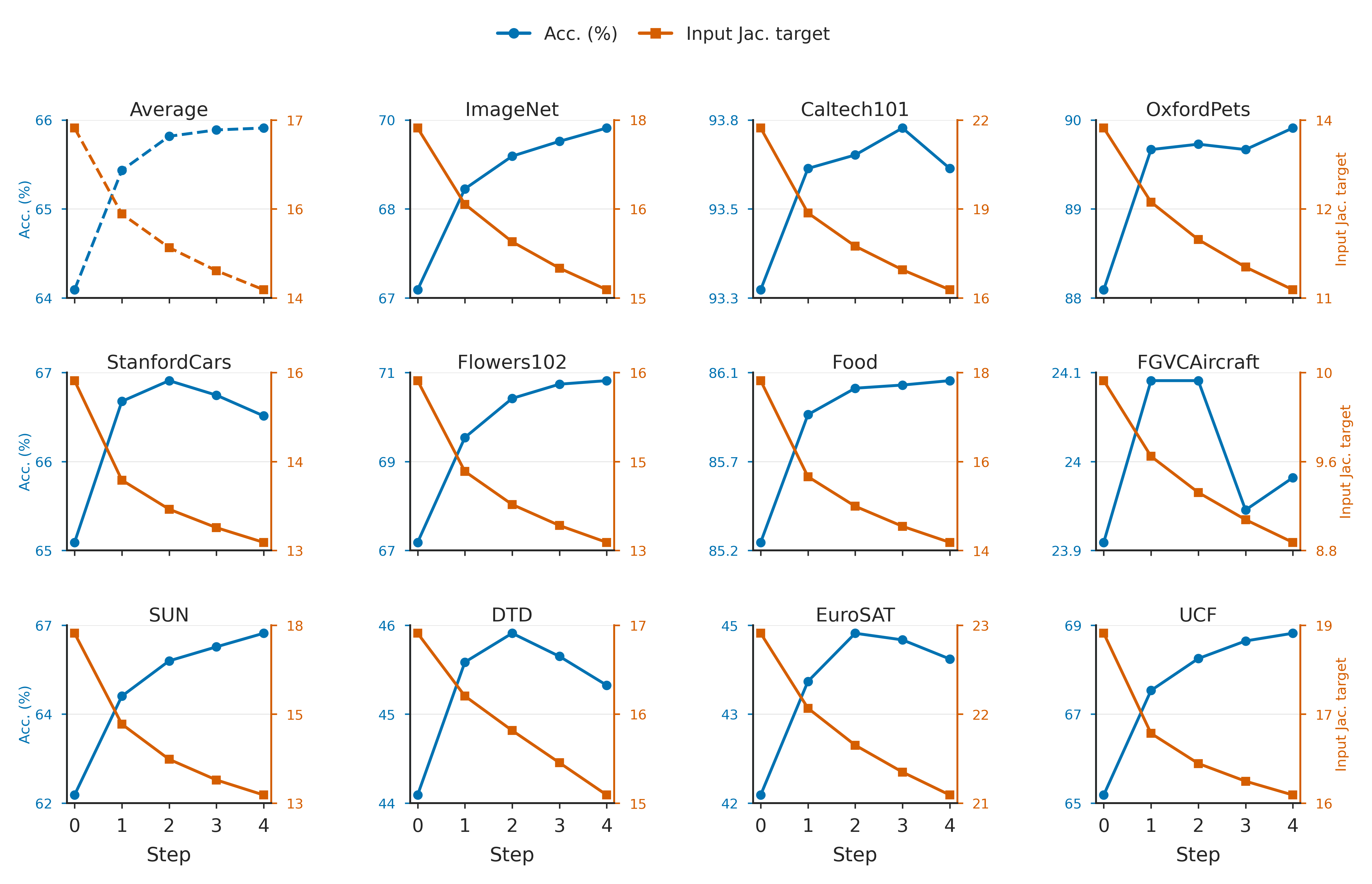}
    \caption{Accuracy and Jacobian norm of the target logit with respect to the input in the cross-dataset evaluation settings. We report each metric for every dataset and across varying numbers of reasoning steps.}
    \label{fig:accuracy-input-jacobian-target-xd-per-dataset}
\end{figure}
\begin{figure}[htbp]
    \centering
    \includegraphics[width=0.95\textwidth]{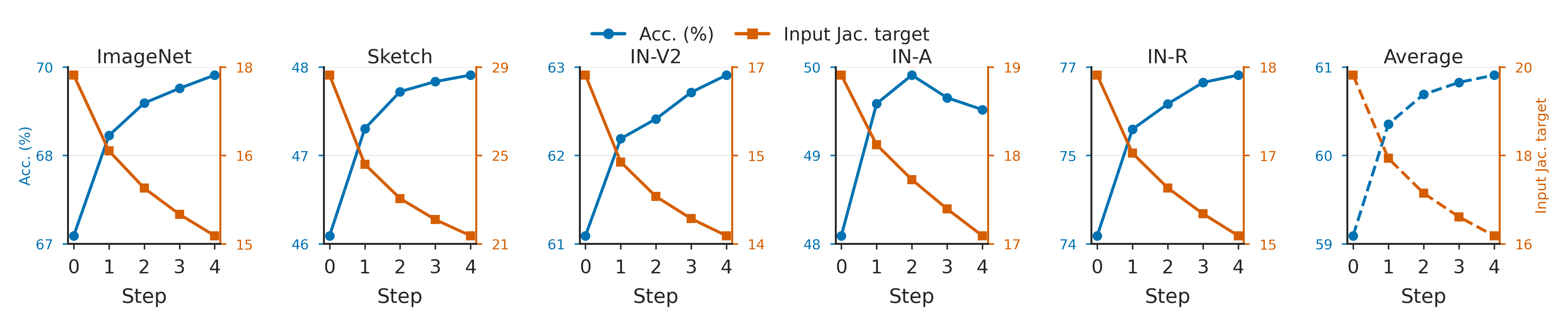}
    \caption{Accuracy and Jacobian norm of the target logit with respect to the input in the out-of-distribution evaluation settings. We report each metric for every dataset and across varying numbers of reasoning steps.}
    \label{fig:accuracy-input-jacobian-target-ood-per-dataset}
\end{figure}
\begin{figure}[htbp]
    \centering
    \includegraphics[width=0.95\textwidth]{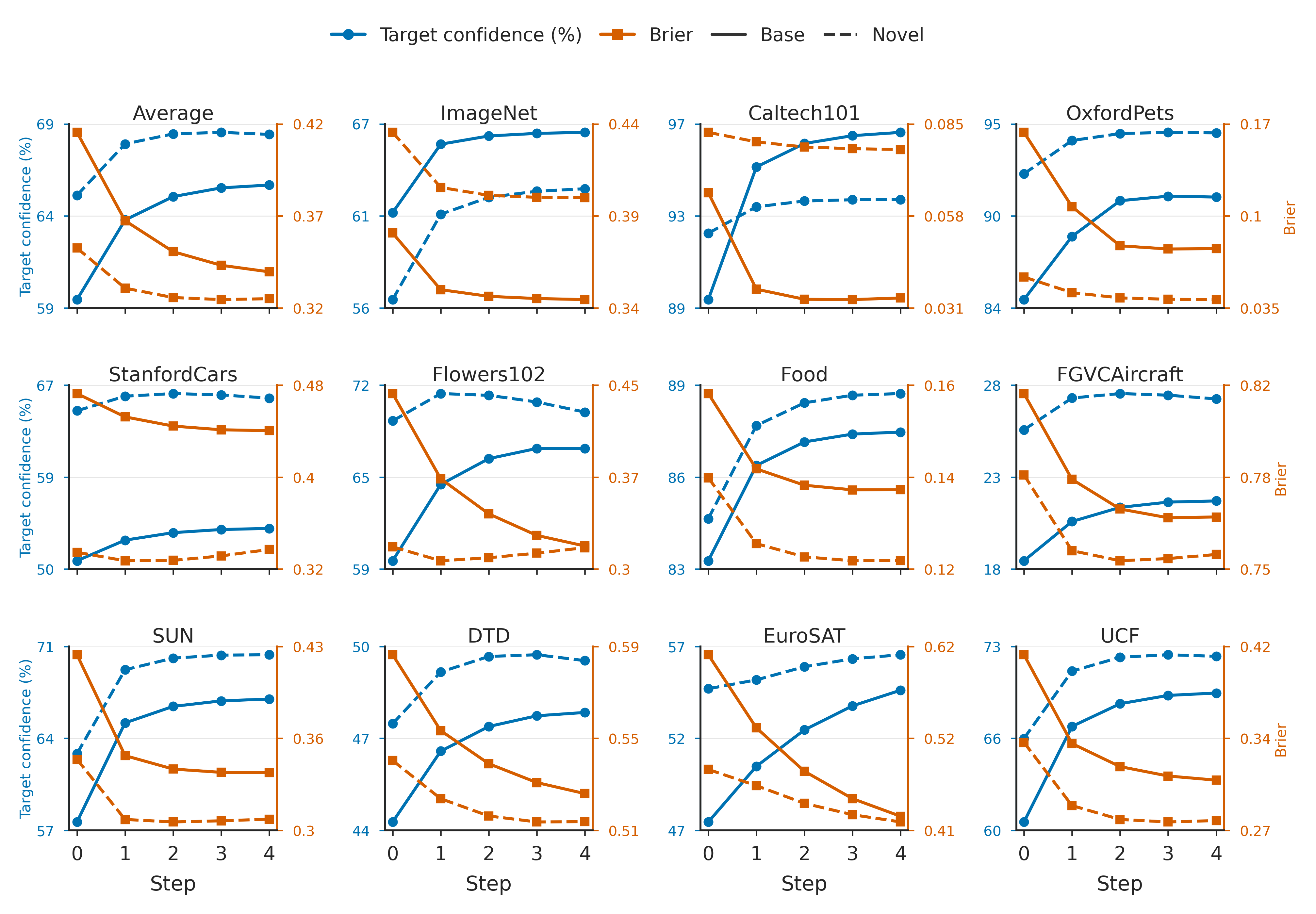}
    \caption{Target confidence and Brier score in the base-to-novel evaluation settings. We report each metric for every dataset and across varying numbers of reasoning steps.}
    \label{fig:target-conf-brier-b2n-per-dataset}
\end{figure}
\begin{figure}[htbp]
    \centering
    \includegraphics[width=0.95\textwidth]{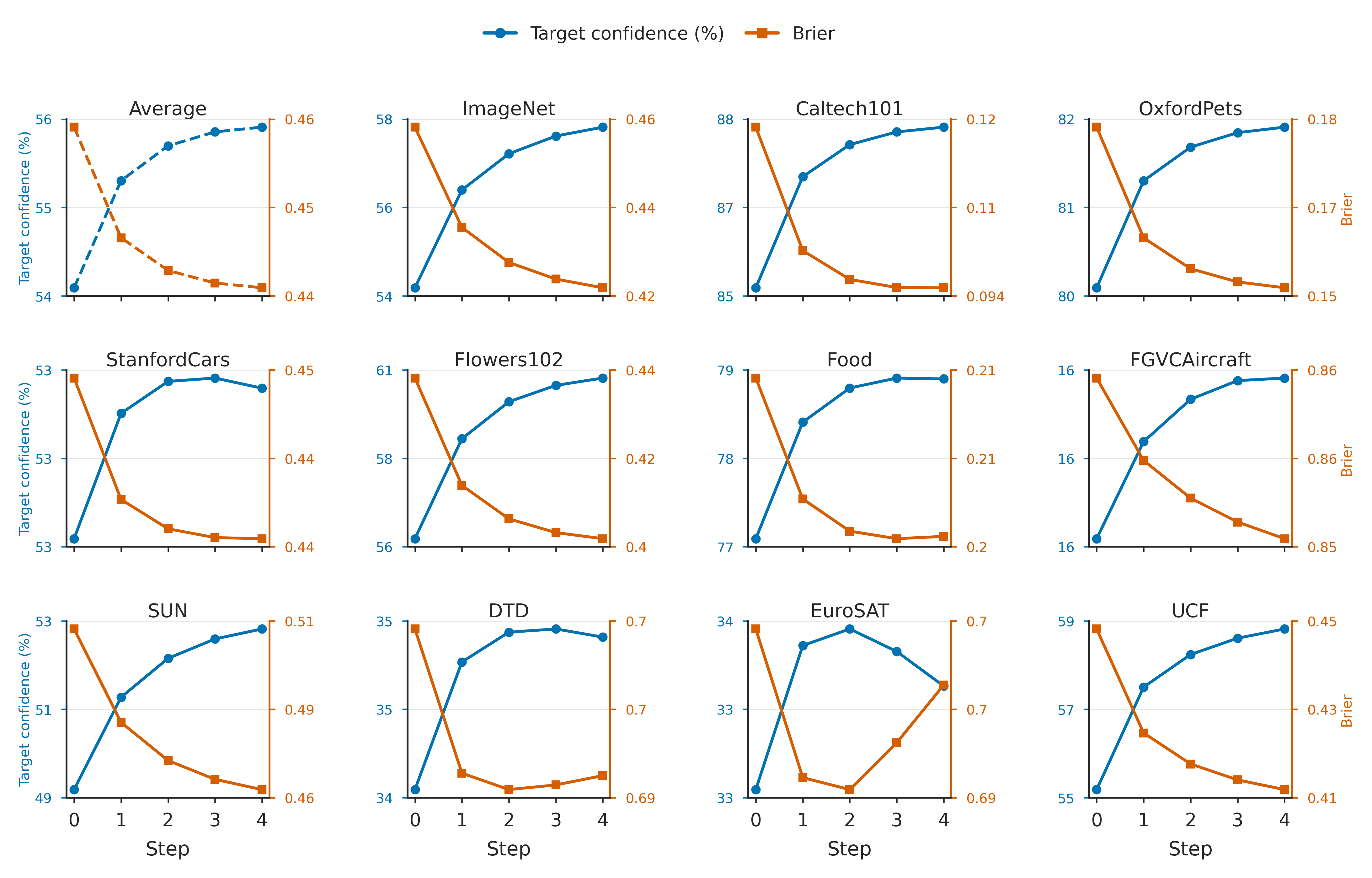}
    \caption{Target confidence and Brier score in the cross-dataset evaluation settings. We report each metric for every dataset and across varying numbers of reasoning steps.}
    \label{fig:target-conf-brier-xd-per-dataset}
\end{figure}
\begin{figure}[htbp]
    \centering
    \includegraphics[width=0.95\textwidth]{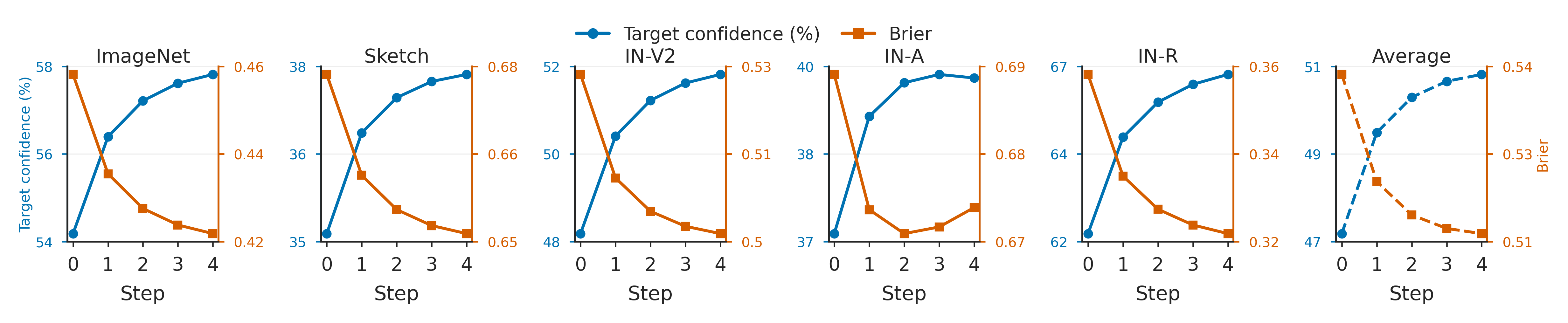}
    \caption{Target confidence and Brier score in the out-of-distribution evaluation settings. We report each metric for every dataset and across varying numbers of reasoning steps.}
    \label{fig:target-conf-brier-ood-per-dataset}
\end{figure}

We extend the evaluation of Section ~\ref{sec:dynamics} to include single datasets in base-to-novel (Figs. ~\ref{fig:target-conf-brier-b2n-per-dataset}, ~\ref{fig:accuracy-input-jacobian-target-b2n-per-dataset}), cross-dataset (Figs. ~\ref{fig:target-conf-brier-xd-per-dataset}, ~\ref{fig:accuracy-input-jacobian-target-xd-per-dataset}) and out-of-distribution evaluation (Figs. ~\ref{fig:target-conf-brier-ood-per-dataset}, ~\ref{fig:accuracy-input-jacobian-target-ood-per-dataset}). 

\begin{figure}[htbp]
    
    \centering
    \includegraphics[width=0.95\textwidth]{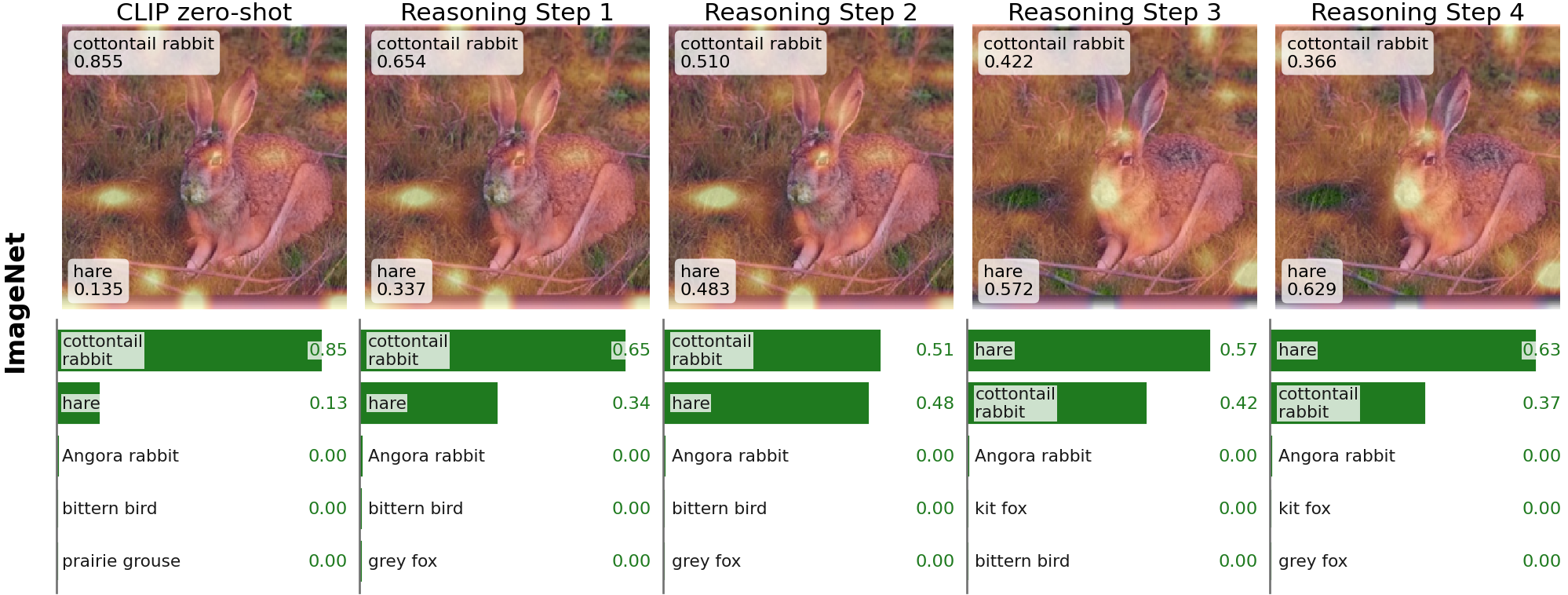}
    \includegraphics[width=0.95\textwidth]{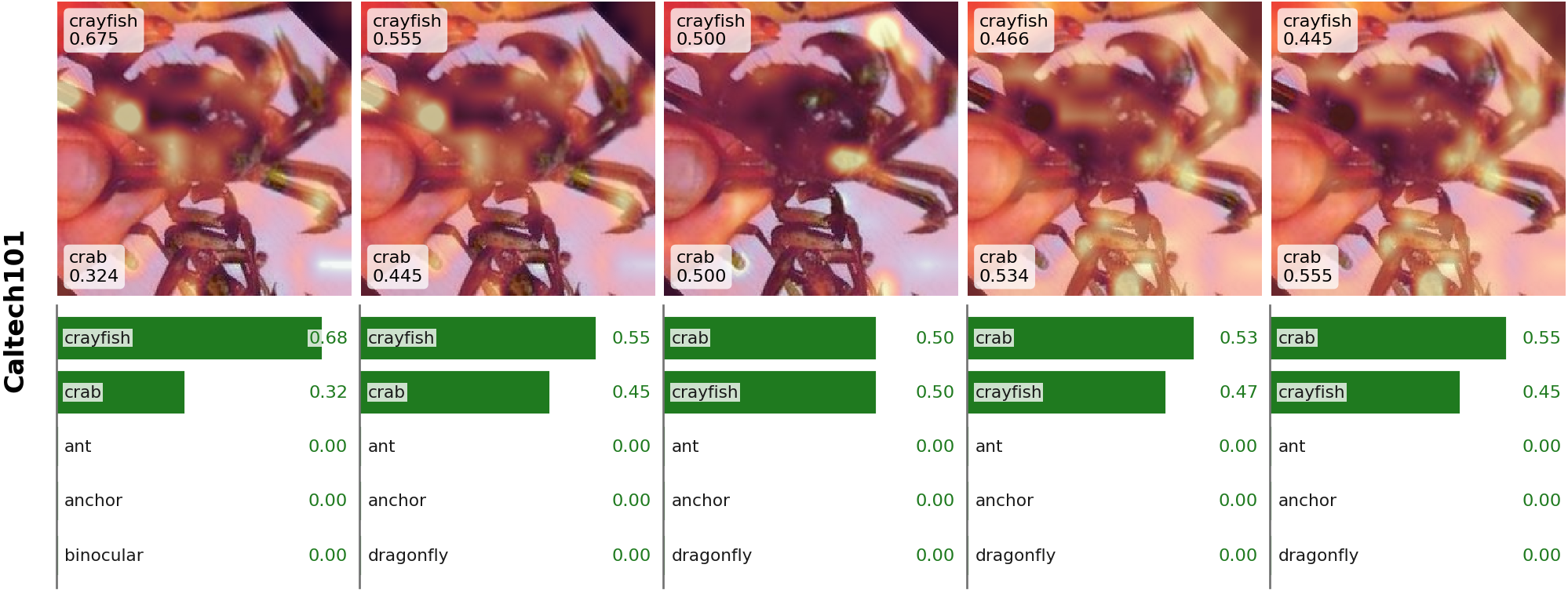}
    \includegraphics[width=0.95\textwidth]{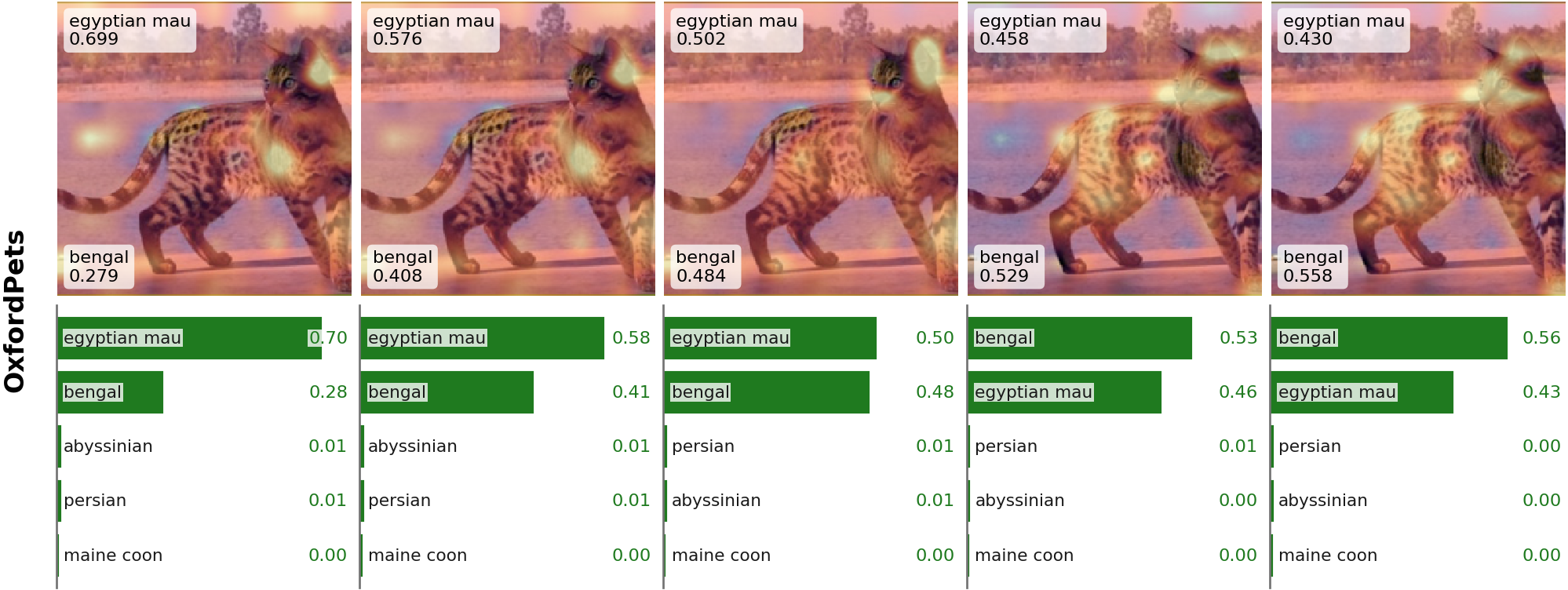}
    \includegraphics[width=0.95\textwidth]{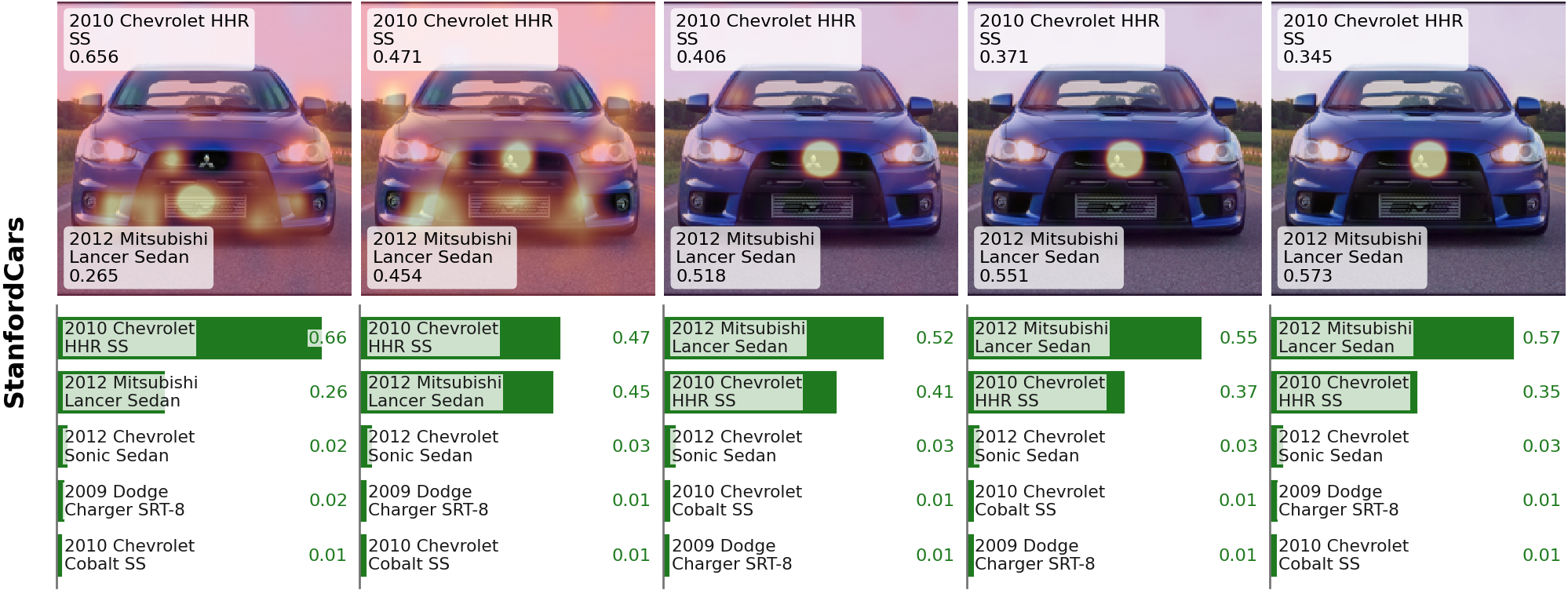}
    \caption{\textbf{Examples of wrong-to-correct refinements through reasoning.} The model progressively shifts its confidence from an incorrect initial prediction to the correct class across reasoning steps.}
    \label{fig:image_grid_full0}
\end{figure}

\begin{figure}[htbp]
    
    \centering
    
    \includegraphics[width=0.95\textwidth]{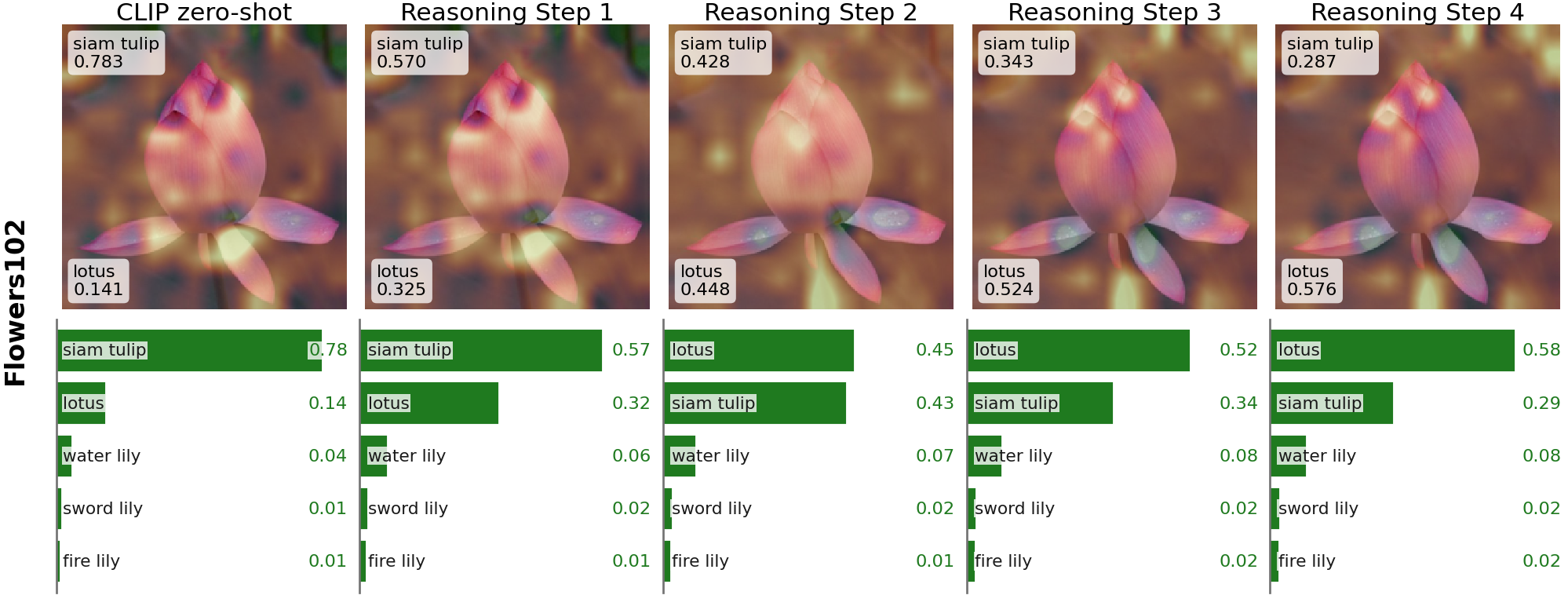}
    \includegraphics[width=0.95\textwidth]{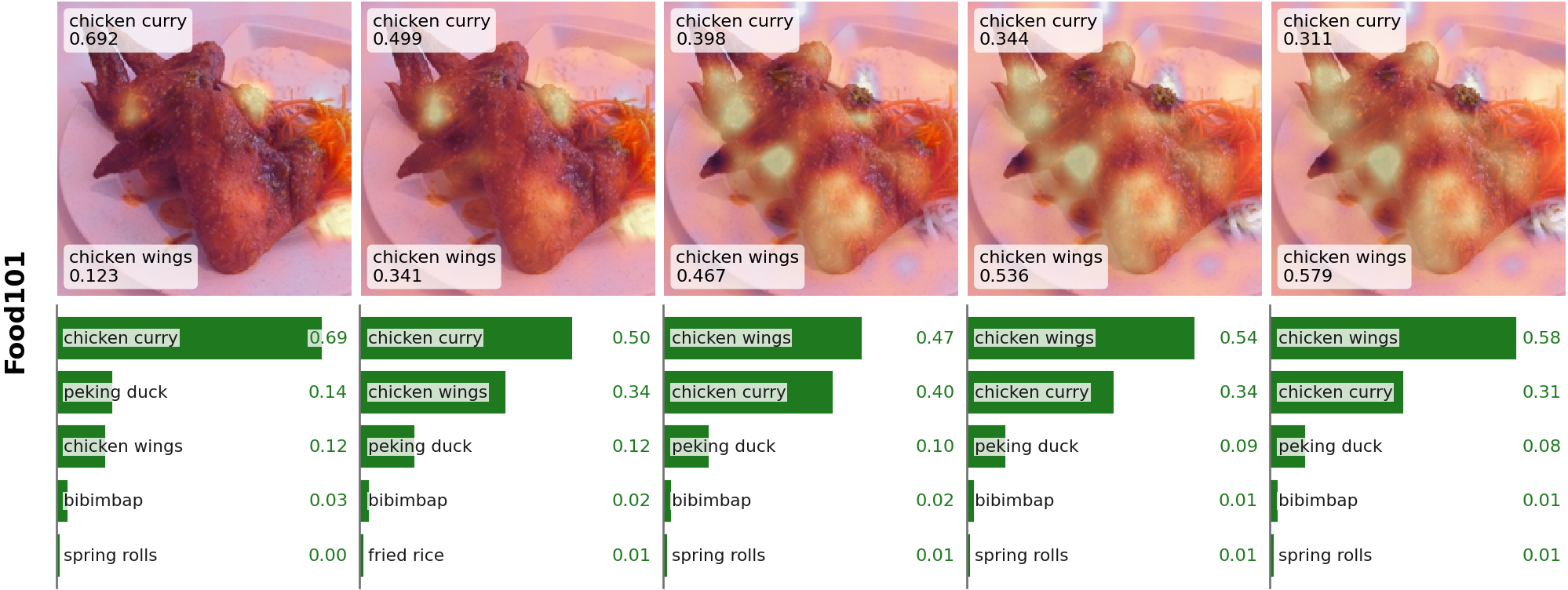}
    \includegraphics[width=0.95\textwidth]{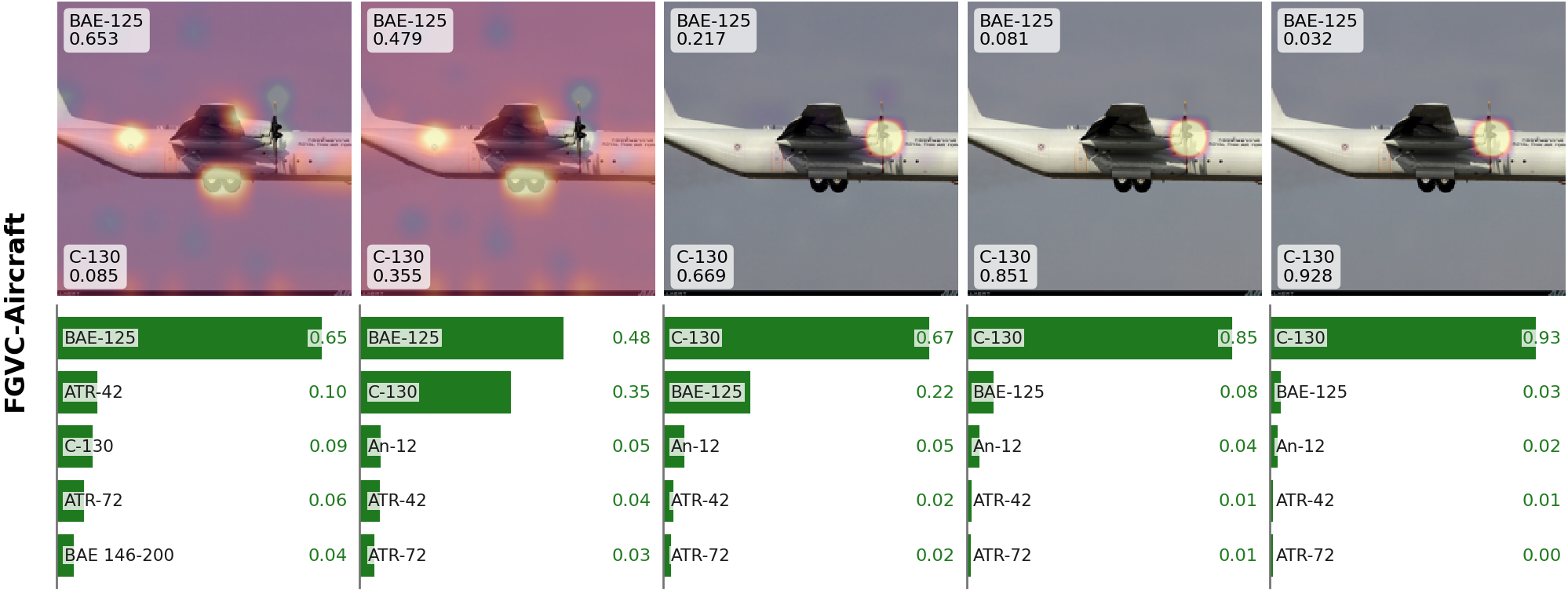}
    \includegraphics[width=0.95\textwidth]{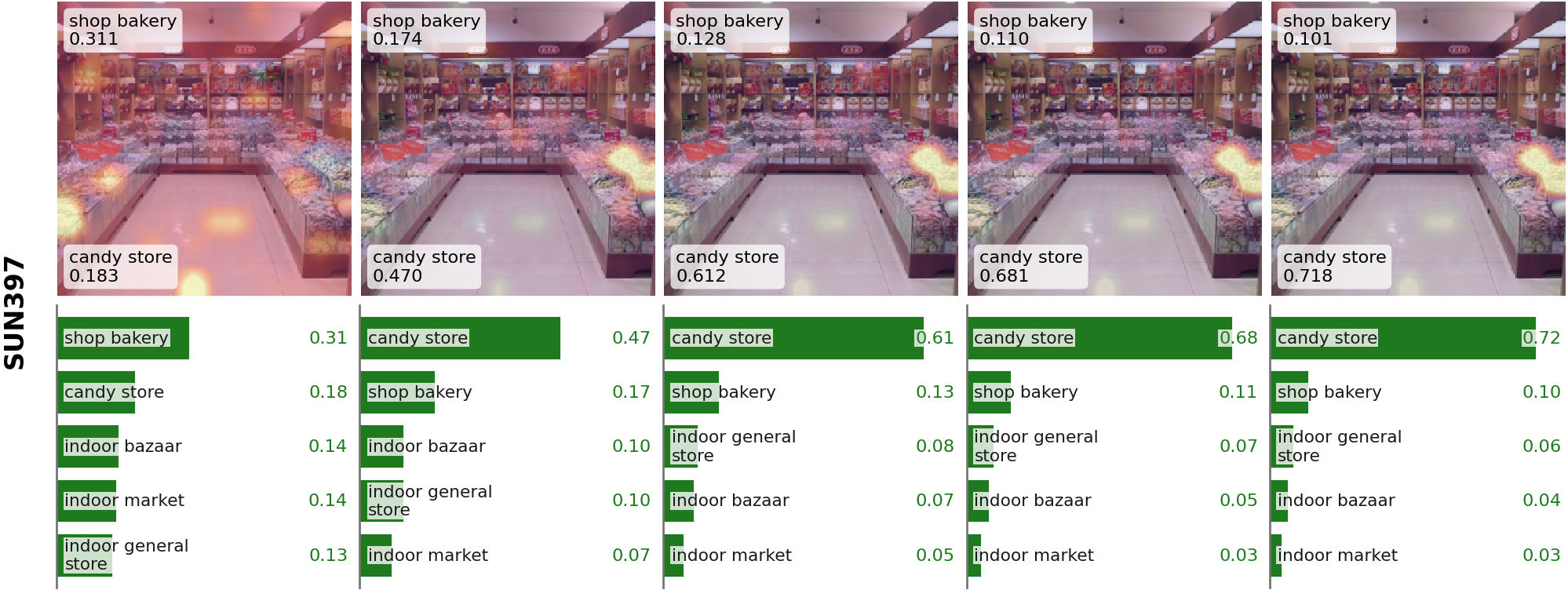}
    \caption{\textbf{Examples of wrong-to-correct refinements through reasoning.} The model progressively shifts its confidence from an incorrect initial prediction to the correct class across reasoning steps.}
    \label{fig:image_grid_full1}
    
\end{figure}

\begin{figure}[htbp]
    
    \centering
    
    \includegraphics[width=0.95\textwidth]{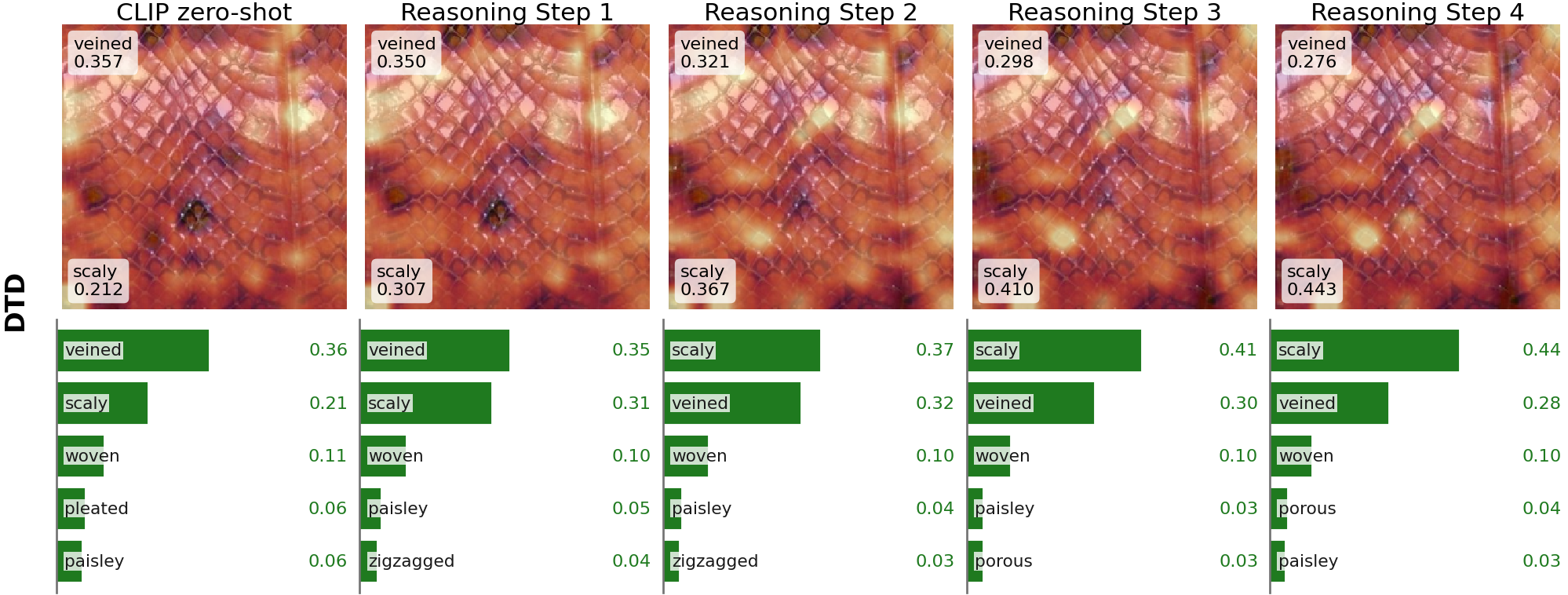}
    \includegraphics[width=0.95\textwidth]{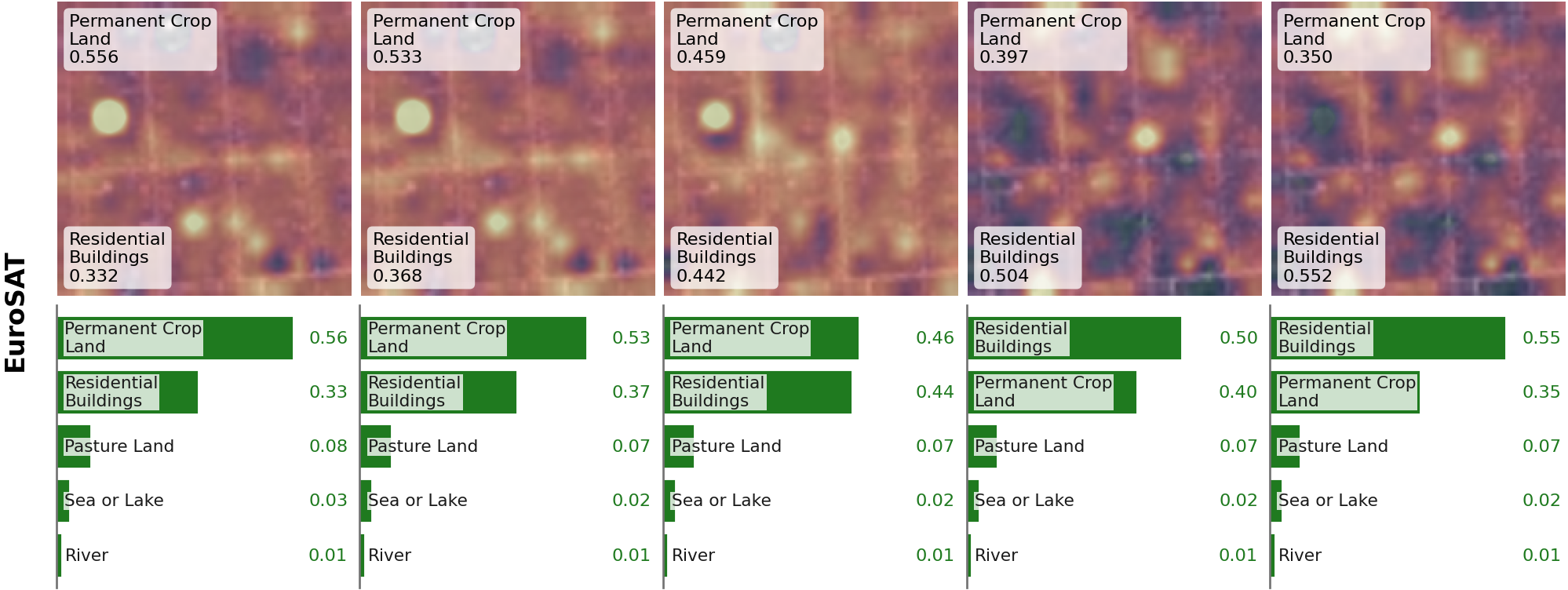}
    \includegraphics[width=0.95\textwidth]{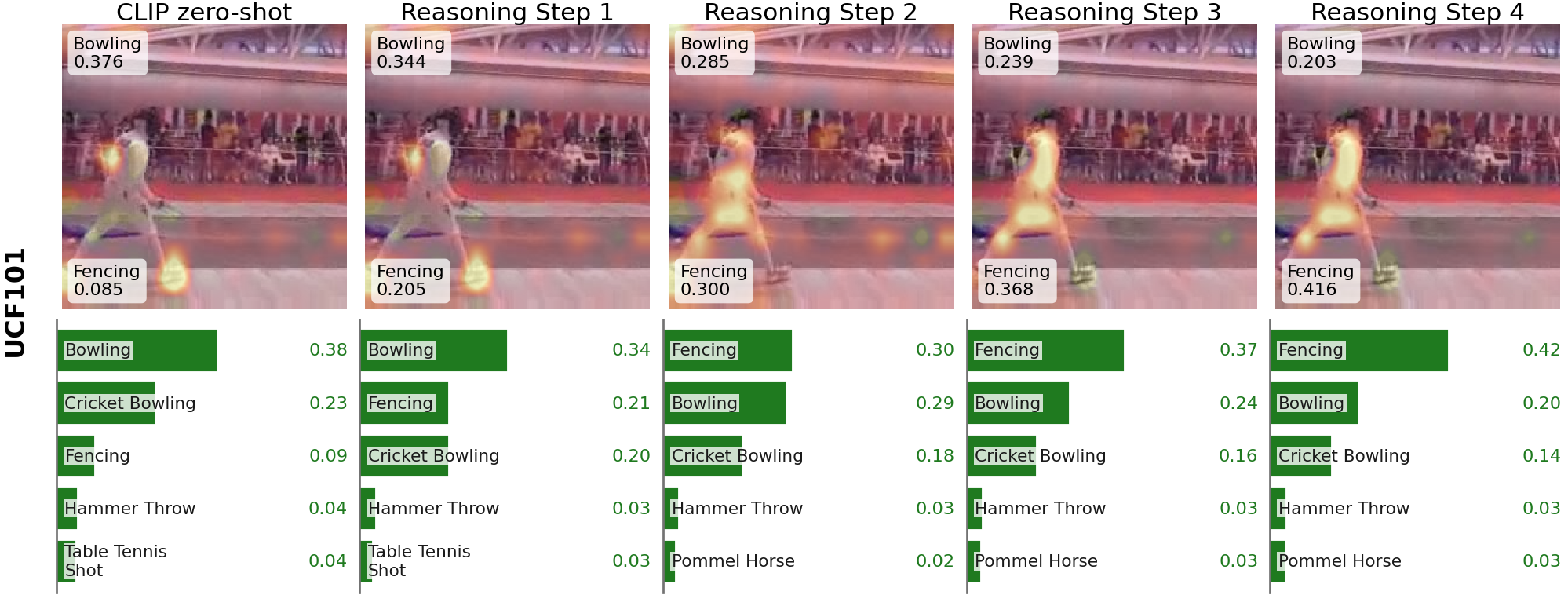}
    \includegraphics[width=0.95\textwidth]{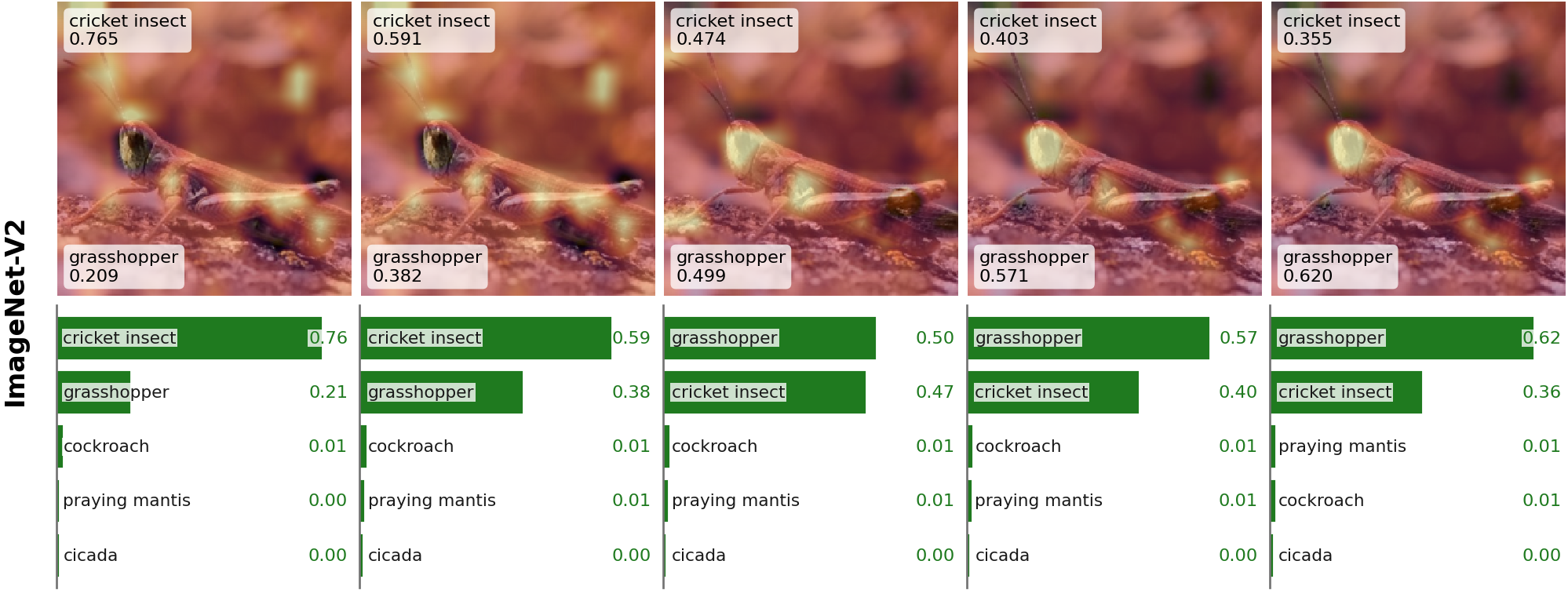}
    \caption{\textbf{Examples of wrong-to-correct refinements through reasoning.} The model progressively shifts its confidence from an incorrect initial prediction to the correct class across reasoning steps.}
    \label{fig:image_grid_full2}
    
\end{figure}

\begin{figure}[htbp]
    
    \centering
    \includegraphics[width=0.95\textwidth]{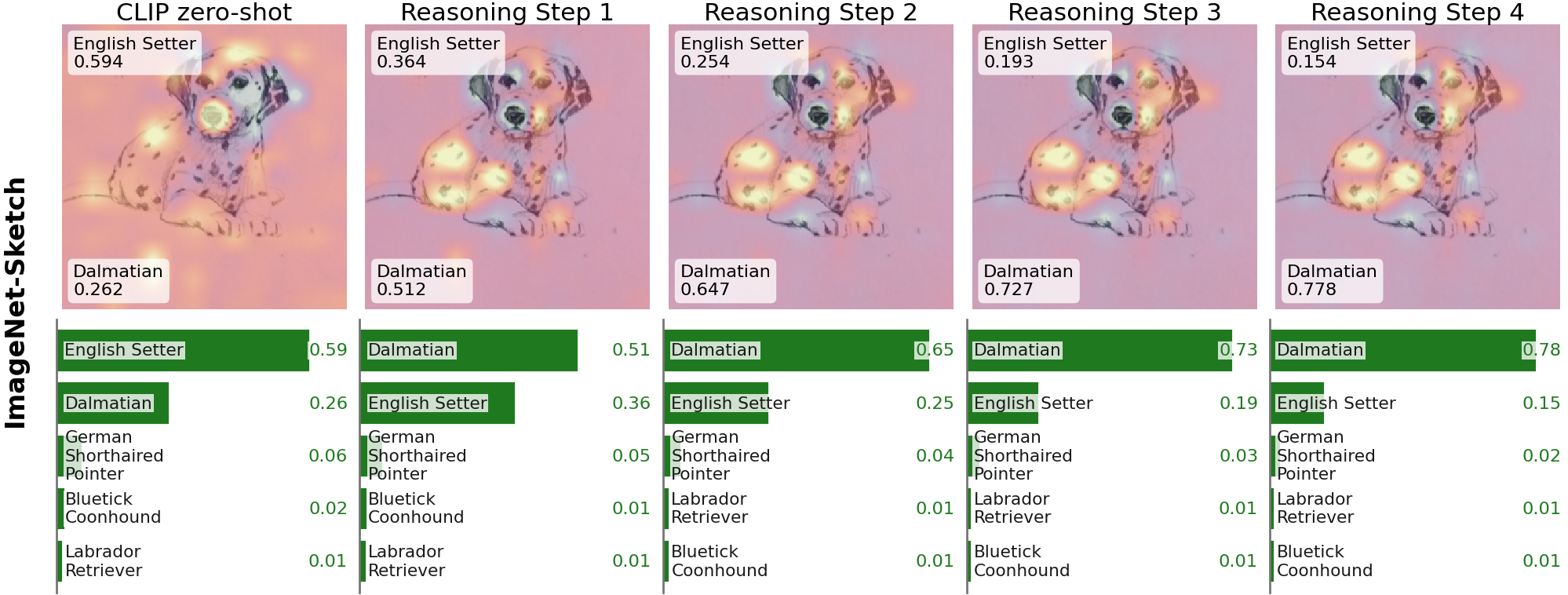}
    \includegraphics[width=0.95\textwidth]{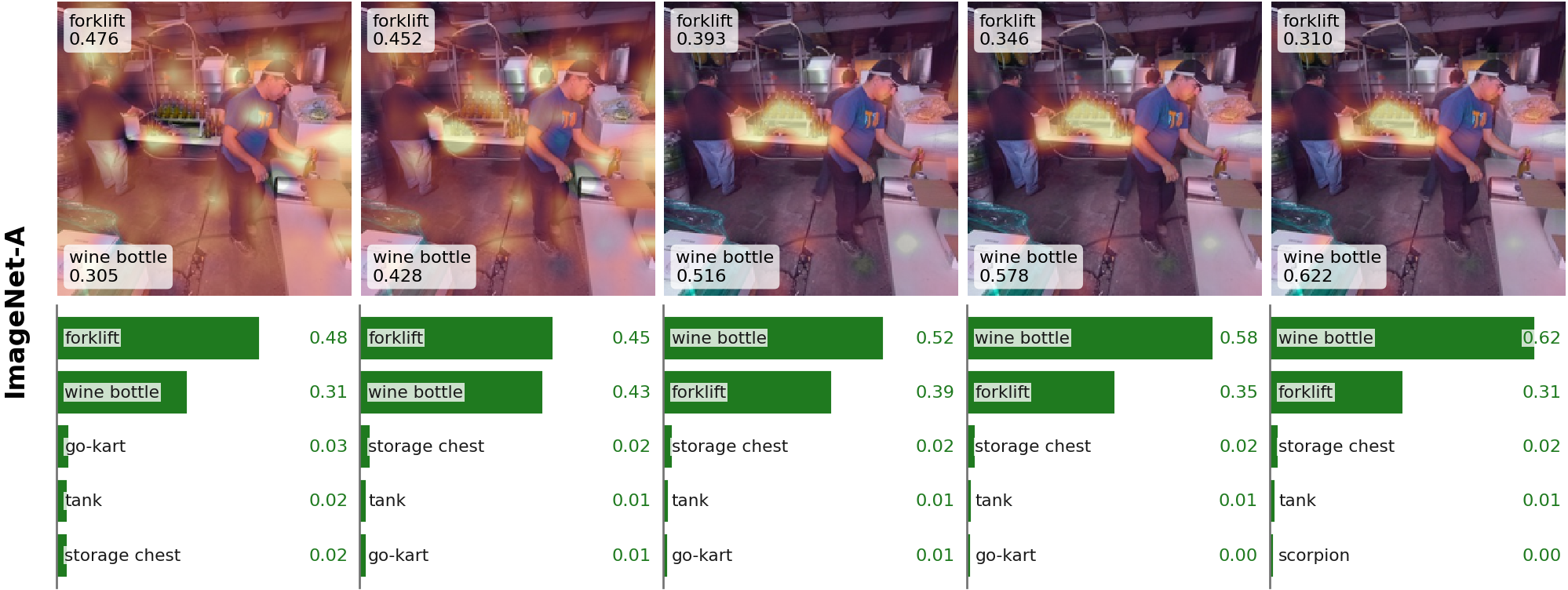}
    \includegraphics[width=0.95\textwidth]{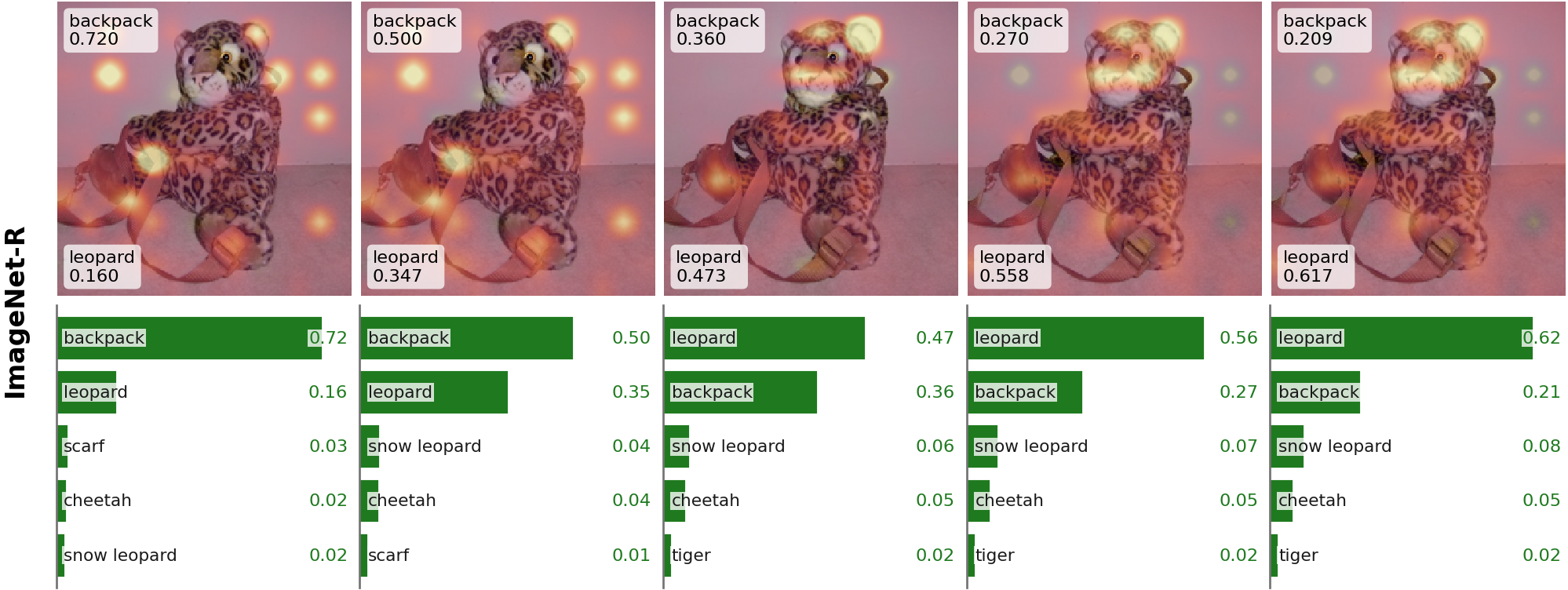}
    \caption{\textbf{Examples of wrong-to-correct refinements through reasoning.} The model progressively shifts its confidence from an incorrect initial prediction to the correct class across reasoning steps.}
    \label{fig:image_grid_full3}
    
\end{figure}

We also show additional visual evidence of reasoning steps across datasets inf Figs. 
~\ref{fig:image_grid_full0}, ~\ref{fig:image_grid_full1}, ~\ref{fig:image_grid_full2}, ~\ref{fig:image_grid_full3}). 

\begin{figure}[htbp]
    \centering
    \includegraphics[width=0.95\textwidth]{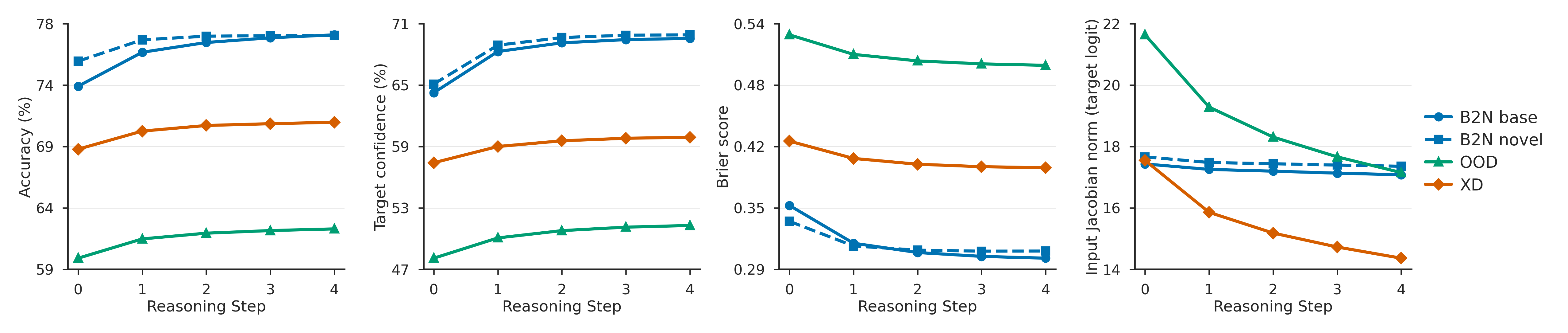}
    \caption{Evaluation metrics across varying numbers of reasoning steps in the three settings: base-to-novel, cross-dataset, and out-of-distribution. We report accuracy, target confidence, Brier score, and Jacobian norm of target logit with respect to input.
}
    \label{fig:accuracy-target-conf-brier-input-jacobian-target-allsettings}
\end{figure}

